\documentclass[journal]{IEEEtran}

\usepackage{cite}

\ifCLASSINFOpdf
\usepackage[pdftex]{graphicx}
   \graphicspath{{../pdf/}{../jpeg/}}
   \DeclareGraphicsExtensions{.pdf,.jpeg,.png}
\else
\fi

\usepackage{booktabs}
\usepackage{mathrsfs}
\usepackage{subfigure}
\usepackage{graphicx}
\usepackage{verbatim}
\usepackage{arydshln}

\usepackage{color}
\usepackage[table]{xcolor} 
\usepackage{multirow}
\usepackage{makecell}
\usepackage{amsmath}
\usepackage{pifont}
\usepackage{amsfonts,amssymb} 
\usepackage{algorithmic}

\usepackage{array}

\usepackage{tikz}
\newcommand*\circled[1]{\tikz[baseline=(char.base)]{\node[shape=circle,draw,inner sep=0.5pt] (char) {\scriptsize#1};}}

\ifCLASSOPTIONcaptionsoff
  \usepackage[nomarkers]{endfloat}
 \let\MYoriglatexcaption\caption
\renewcommand{\caption}[2][\relax]{\MYoriglatexcaption[#2]{#2}}
\fi

\DeclareMathOperator{\Softmax}{Softmax}
\DeclareMathOperator{\CvT}{CvT}
\DeclareMathOperator{\GAP}{GAP}
\DeclareMathOperator{\MLP}{MLP}
\DeclareMathOperator{\ConV}{ConV}
\DeclareMathOperator{\ExP}{ExP}
\DeclareMathOperator{\ReLU}{ReLU}
\DeclareMathOperator{\FCfirst}{FC_1}
\DeclareMathOperator{\FCsecond}{FC_2}
\DeclareMathOperator{\Sigmoid}{Sigmoid}

\usepackage{hyperref}
\hypersetup{
	colorlinks=true,
	linkcolor=blue,
	citecolor = blue,      
}

\usepackage{diagbox}

\begin{document}

\title{Exploring Rich Subjective Quality Information for Image Quality Assessment in the Wild}
\author{Xiongkuo Min,~\IEEEmembership{Member,~IEEE}, Yixuan Gao, Yuqin Cao, Guangtao Zhai,~\IEEEmembership{Senior Member,~IEEE},\\\hspace{0.55cm}Wenjun Zhang,~\IEEEmembership{Fellow,~IEEE}, Huifang Sun,~\IEEEmembership{Fellow,~IEEE}, and Chang Wen Chen,~\IEEEmembership{Fellow,~IEEE}
 \thanks{X. Min, Y. Gao, Y. Cao, G. Zhai, and W. Zhang are with the Institute of Image Communication and Network Engineering, Shanghai Jiao Tong University, Shanghai 200240, China (e-mail: minxiongkuo@sjtu.edu.cn; gaoyixuan@sjtu.edu.cn; caoyuqin@sjtu.edu.cn; zhaiguangtao@sjtu.edu.cn; zhangwenjun@sjtu.edu.cn).}
\thanks{H. Sun is with the Peng Cheng Laboratory (PCL), Shenzhen 518055, China (e-mail: hsun138@yahoo.com).}
\thanks{C. W. Chen is with The Hong Kong Polytechnic University, Hong Kong (e-mail: changwen.chen@polyu.edu.hk).}
\vspace{-0.1cm}
}

\maketitle 

\begin{abstract}
Traditional in the wild image quality assessment (IQA) models are generally trained with the quality labels of \textit{mean opinion score} (MOS), while missing the rich subjective quality information contained in the quality ratings, for example, the \textit{standard deviation of opinion scores} (SOS) or even \textit{distribution of opinion scores} (DOS). In this paper, we propose a novel IQA method named \underline{\smash{RichIQA}} to explore the rich subjective rating information beyond MOS to predict image quality in the wild. RichIQA is characterized by two key novel designs: \circled{1} a three-stage image quality prediction network which exploits the powerful feature representation capability of the Convolutional vision Transformer (CvT) and mimics the short-term and long-term memory mechanisms of human brain; \circled{2} a multi-label training strategy in which rich subjective quality information like MOS, SOS and DOS are concurrently used to train the quality prediction network. 
Powered by these two novel designs, RichIQA is able to predict the image quality in terms of a distribution, from which the mean image quality can be subsequently obtained. 
Extensive experimental results verify that the three-stage network is tailored to predict rich quality information, while the multi-label training strategy can fully exploit the potentials within subjective quality rating and enhance the prediction performance and generalizability of the network. 
RichIQA outperforms state-of-the-art competitors on multiple large-scale in the wild IQA databases with rich subjective rating labels. The code of RichIQA will be made publicly available on GitHub.
\end{abstract}

\begin{IEEEkeywords}
Image quality assessment, in the wild, mean opinion score, opinion score distribution, multi-label training strategy, three-stage network.
\end{IEEEkeywords}

%
\IEEEpeerreviewmaketitle

\section{Introduction}
\label{sec:intro}
\IEEEPARstart Image quality assessment (IQA) aims to evaluate the perceptual quality of images through subjective or objective evaluation methods \cite{zhai2020perceptual,min2024perceptual,lin2011perceptual,min2021screen}.  
Subjective IQA collects subjective opinion scores by inviting a large number of observers to rate the visual quality of images, whose procedures have been standardized by some international organizations like the International Telecommunication Union (ITU) \cite{ITUbt500,ITUp910}. Objective IQA aims to develop computation models capable of predicting perceptual image quality in a manner that correlates well with human perception. 
Objective IQA can be classified into three categories: full-reference (FR) IQA \cite{zhang2018unreasonable,bosse2017deep,gu2017fast,min2020study}, reduced-reference (RR) IQA \cite{wang2016reduced,yu2022perceptual,huang2020perceptual}, and no-reference (NR) IQA \cite{zhou2024multitask,li2021blind}. Both FR and RR IQA require information from the reference image to evaluate the quality of the distorted image. In practice, the acquisition of the reference is challenging, making NR IQA the predominant method for the IQA task.

Early NR IQA methods focus primarily on evaluating the quality of synthetically distorted images. Synthetically distorted images are generated by artificially adding various degradations such as compression, blur, noise, and color distortions. Typical IQA databases for synthetic distortions include LIVE \cite{livedatabase}, CSIQ \cite{larson2010most}, TID2013 \cite{ponomarenko2015image}, VCLFER \cite{zaric2011vcl}, and so on. 
The images in these databases have relatively simple distortions, allowing most NR IQA methods to achieve good performance on synthetic distortion databases \cite{mittal2012no,mittal2012making,saad2012blind}. 
However, authentically distorted images in the wild are more commonly encountered in practical applications, which are affected by distortions arising from various sources, such as sensor noise, transmission errors, and compression. These distortions are more complex and diverse, and they usually mix with each other. Representative in the wild IQA databases include KonIQ-10K \cite{2019KonIQ}, BID \cite{ciancio2010no}, LIVE Challenge \cite{ghadiyaram2015massive}, SPAQ \cite{fang2020perceptual}, FLIVE \cite{ying2020patches}, and so on. 
Though traditional NR IQA methods can be also used to assess the quality of in the wild images, their performances are usually not the best \cite{song2022blind}. Therefore, it is important to develop IQA methods specifically tailored to images in the wild to ensure accurate evaluations.

Extracting quality-aware image features has long been the core of designing quality measures for images in the wild. Traditional IQA methods extract low-level handcrafted image features, which have limited effectiveness in predicting the quality of in the wild images. In contrast, convolution neural network (CNN) based IQA methods, which can extract both low-level and high-level image features, have been proven effective for in the wild IQA. For example, DBCNN \cite{zhang2018blind} demonstrates good prediction performance on the LIVE Challenge database. TReS \cite{golestaneh2021no} and DACNN \cite{pan2022dacnn} are also effective on IQA databases in the wild. 
Furthermore, researchers have developed IQA methods that are specifically designed for images in the wild. For example, Su \textit{et al}. introduced an IQA method tailored for images in the wild using a hyper network \cite{su2020blindly}. Li \textit{et al}. proposed a coarse-to-fine IQA method for images in the wild \cite{li2024reqa}. Sun \textit{et al}. developed an IQA method that leverages hierarchical feature fusion and iterative mixed database training, which demonstrates the superiority of using multi-level features for predicting image quality \cite{sun2023blind}.

In recent years, the performances of IQA in the wild have been boosted by a large margin by designing novel network architectures and utilizing cutting-edge learning paradigms, however only marginal performance gains can be obtained by following this technical route nowadays. 
Besides the network design, potential gain can also originate from the subjective quality rating data, which has rarely been explored in the current IQA literatures. 
Through subjective IQA, a group of ratings (typically more than 15 ratings according to ITU-R BT.500 \cite{ITUbt500}) are collected for each image, from which the \textit{mean opinion score} (\textbf{MOS}) is calculated as image quality. Since the MOS is deemed as the ground-truth of image quality, MOSs are usually used as training labels and prediction targets in most existing IQA models. 
However, the raw quality ratings contain rich subjective information which cannot be described by a single MOS, especially for images in the wild where different subjects can have larger divergences.

\begin{figure}[t]
	\centering
	\includegraphics[width=3.48in]{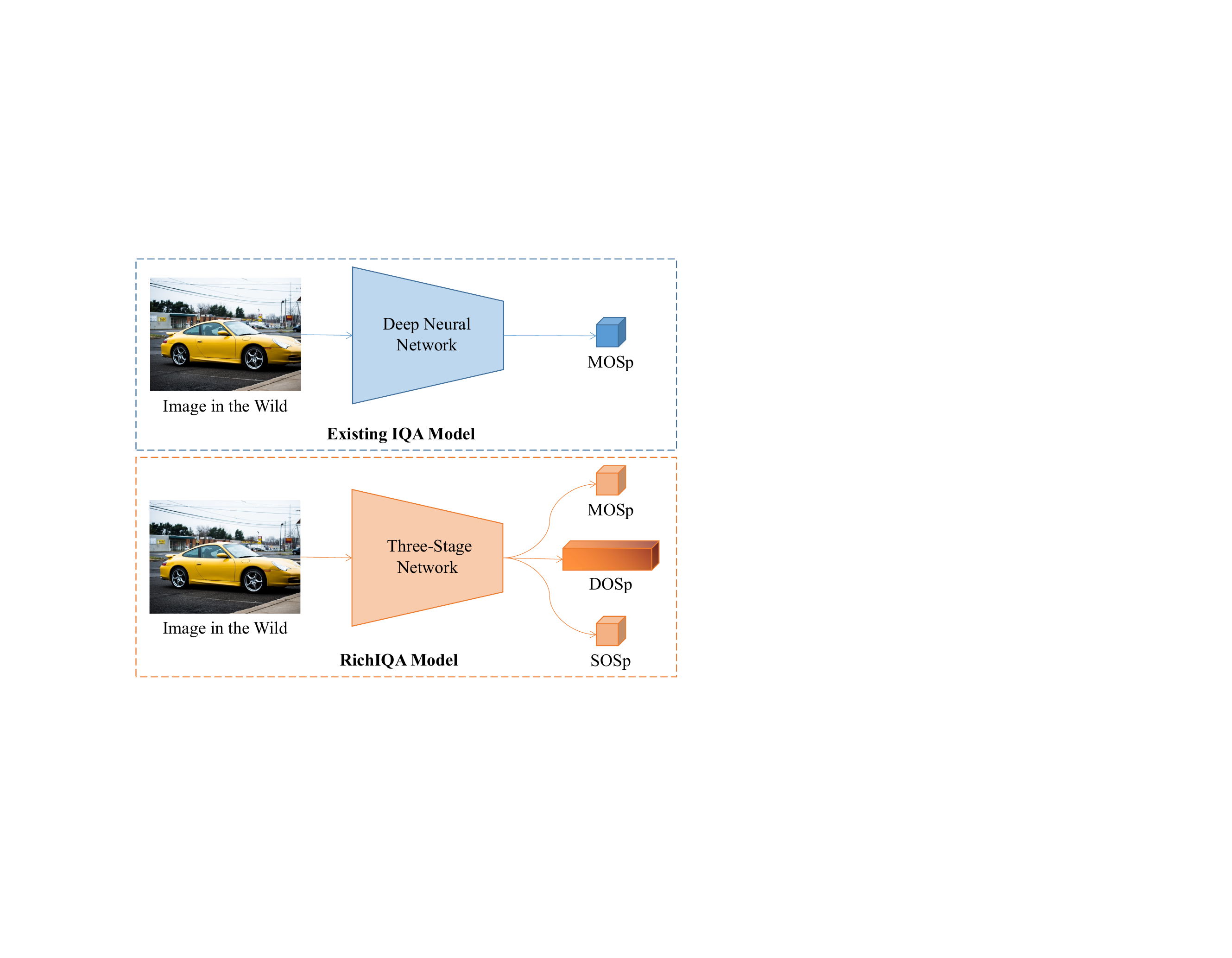}
	\caption{Existing IQA models vs. RichIQA model. Existing IQA models predict image quality generally described by MOS, while RichIQA explores rich subjective quality rating information and predicts image quality described by MOS, DOS, and SOS (denoted as MOSp, DOSp, SOSp, where the postfix `p' indicates `predicted').}
	\label{fig_comparison}
 \vspace{-10pt}
\end{figure}

Besides MOS, various other statistical measures of the quality ratings also provide rich subjective information about image quality, for example, the \textit{standard deviation of opinion scores} (\textbf{SOS}), and the \textit{distribution of opinion scores} (\textbf{DOS}). Among these measures, MOS provides the average level of image quality \cite{ITUbt500,ITUp910}; SOS describes the dispersion of image quality scores reflecting subjective diversity \cite{2011SOS}; DOS provides the probability distribution of image quality across different quality labels, preserving comprehensive subjective quality information \cite{gao2023blind}. 
Such rich subjective quality information is completely missed by the existing IQA models. To fill this void, we design an in the wild IQA model named \textbf{RichIQA} to explore and predict the rich image quality information including MOS, DOS and SOS, which distinguishes RichIQA from the existing IQA models, as illustrated in Fig. \ref{fig_comparison}. 
More specifically, powered by a novel \textit{three-stage quality prediction network} and a novel \textit{multi-label training strategy}, RichIQA can comprehensively explore the hidden subjective quality information beyond MOS, and achieve superior in the wild IQA performances. 

The \textit{three-stage quality prediction network} exploits the powerful feature representation capability of the Convolutional vision Transformer (CvT) \cite{wu2021cvt} and mimics the short-term and long-term memory (SLM) mechanism of human brain to predict image quality through three modules: a multi-level feature extraction module, a SLM module, and a quality prediction module. 
The multi-level feature extraction module uses a multi-stage CvT to extract rich image features,  where convolution operations capture local image features and transformers capture global image features. 
The SLM module first employs a CNN and a graph convolution network (GCN) \cite{li2015gated} to model the short-term and long-term memory mechanisms respectively, then mimics the human brain processing of image features. 
The quality prediction module learns the mapping from image features extracted by the first two modules to the predicted image quality distribution and simultaneously outputs the mean image quality. 

The \textit{multi-label training strategy} exploits rich subjective quality rating information beyond MOS to effectively train the three-stage quality prediction network. Selecting appropriate training labels with respect to image quality is critical to the image quality prediction network training. Most existing IQA methods are only trained with the MOS labels, however Gao \textit{et al}. \cite{gao2022image} and Talebi \textit{et al}. \cite{2017NIMA} have demonstrated the feasibility of using the DOSs of image quality as training labels. 
Since our quality prediction module is capable of simultaneously outputting a distribution and a mean score of image quality, we propose a multi-label training strategy in which the MOS and DOS of image quality are used as training labels to train the network simultaneously. The strategy cooperates well with the quality prediction network especially the last quality prediction module, and learns the mapping from in the wild image features to the DOS of image quality.

Our contributions can be summarized as three-fold:
\begin{enumerate}
\item We design a novel IQA network RichIQA that integrates the multi-stage CvT with the human brain's SLM mechanism to predict the quality of in the wild images.
\item To enable the designed network to learn more subjective quality information of images, we propose a multi-label training strategy, which improves the prediction performance and generalizability of the network.
\item Experimental results indicate that RichIQA, with low computational cost,  outperforms state-of-the-art IQA methods on five in the wild IQA databases.
\end{enumerate}

The rest of this paper is organized as follows. Section \ref{sec:method} details the proposed RichIQA method. Section \ref{sec:result} compares the RichIQA with other state-of-the-art methods. Section \ref{sec:conclusion} concludes the paper.

\begin{figure*}[ht]
\centering
\includegraphics[width=7.1in]{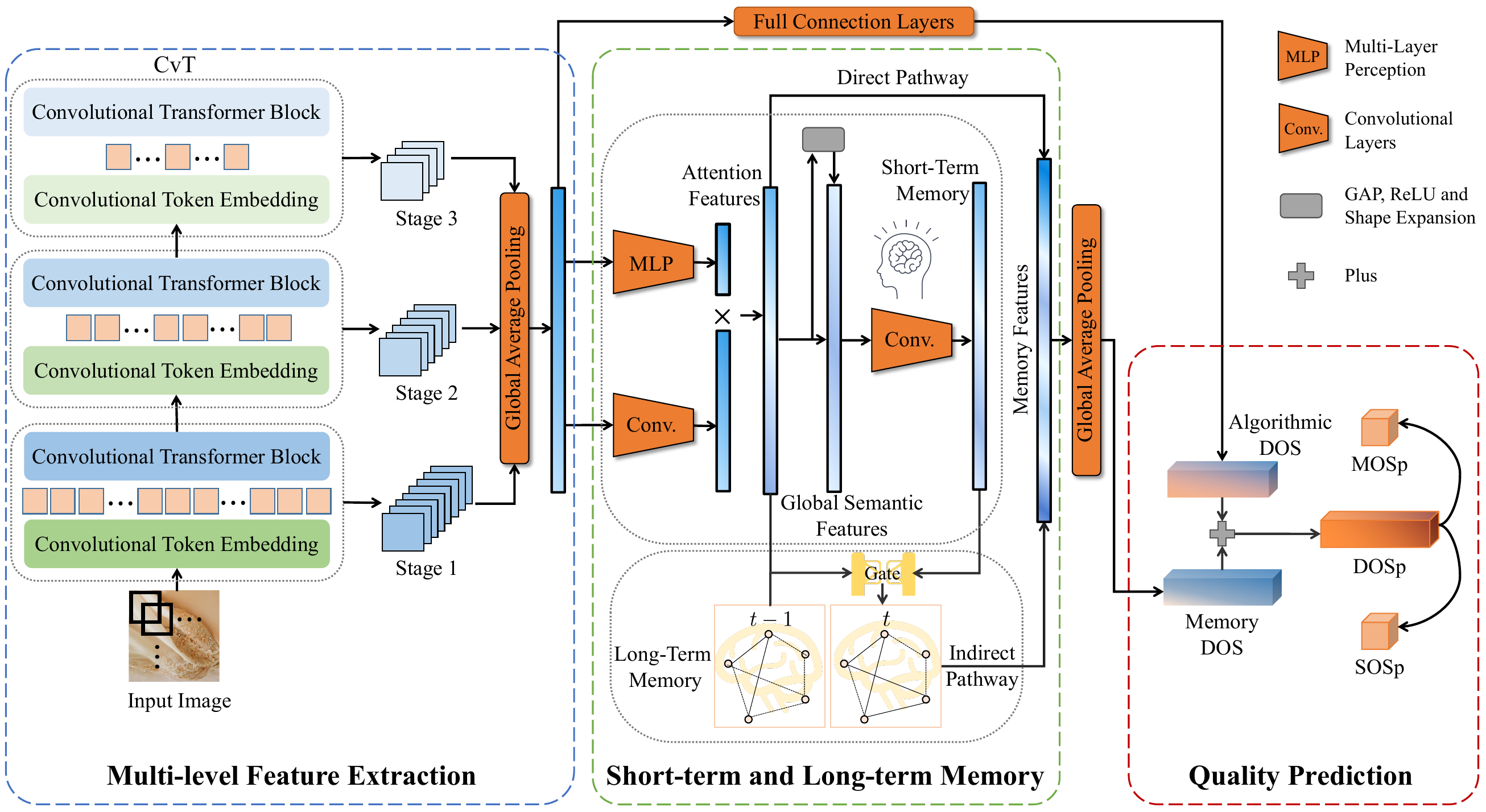}
\caption{Framework of the proposed RichIQA model, which consists of three modules. An image is first input into a multi-level feature extraction module, then the extracted features are inputted into a short-term and long-term memory module, and finally the features extracted from these two modules are used to predict the image quality described by MOS, DOS, and SOS. 
}
 \vspace{-10pt}
\label{framework}
\end{figure*}


\section{Proposed Method}\label{sec:method}
In this section, we provide a detailed description of the proposed  in the wild rich image quality information prediction model \textbf{RichIQA}, which is featured with a \textit{three-stage quality prediction network} and a \textit{multi-label training strategy}. 
The framework of RichIQA is illustrated in Fig. \ref{framework}, whose network consists of three modules: a multi-level feature extraction module, a SLM module, and a quality prediction module. 
To train the three-stage RichIQA network, we make full use of the rich subjective information within the quality ratings, including the MOS, DOS and SOS of image quality. Specifically for IQA databases with different subjective quality labels, we design different training methodologies to improve the applicability of RichIQA to various databases. 

\subsection{Multi-level Feature Extraction}
In traditional Vision Transformer (ViT) \cite{dosovitskiy2020image}, the input image is divided into patches that are processed independently by transformers. A significant limitation of such network is its insufficient exploitation of local information in images, which can be well handled by CNN. To address this issue of ViT, CvT \cite{wu2021cvt} is proposed, a deep learning network that combines the structures of CNN and transformer. Specifically, CvT introduces a convolutional inductive bias into the self-attention mechanism of each transformer. 
This design allows CvT to take into account the relationships between each patch and its surroundings, thereby more effectively capturing local information in images.  In addition, the global self-attention mechanism of the transformers allows CvT to process long-range dependencies in images. 

The multi-level feature extraction of RichIQA is powered by a CvT, from which different levels of image features can be extracted. The multi-stage structure of CvT is particularly beneficial for in the wild image quality modeling, as the perceptual visual quality is affected by both low-level visual features (e.g. distortions) and high-level semantic information (e.g. content) \cite{sun2023blind}. 
For an input image $\mathbf{I}$, RichIQA inputs it into a CvT and extracts different levels of image features from three stages of the CvT:
\begin{align}
   \mathbf{F}_1={\CvT}_1(\mathbf{I}), ~
\mathbf{F}_2={\CvT}_2(\mathbf{F}_1), ~
\mathbf{F}_3={\CvT}_3(\mathbf{F}_2),
\end{align}
where ${\CvT}_i~(i=1,2,3)$ is the $i$-th stage of CvT. $\mathbf{F}_i\in \mathbb{R}^{C_i\times H_i\times W_i} ~(i=1,2,3)$ denotes the features extracted from the $i$-th stage of CvT, in which $C_i$, $H_i$, and $W_i$ represent the channel, height, and width of the feature. 
Then, we input the extracted features $(\mathbf{F}_1, ~\mathbf{F}_2, ~\mathbf{F}_3)$ into a global average pooling (GAP) block and concatenate the pooled features together:
\begin{eqnarray}
\mathbf{F}=\GAP(\mathbf{F}_1)\oplus \GAP(\mathbf{F}_2)\oplus \GAP(\mathbf{F}_3),
\end{eqnarray}
where $\oplus$ is the concatenation operation and $\mathbf{F}\in \mathbb{R}^{C_1+C_2+C_3}$.

\subsection{Short-Term and Long-Term Memory Modeling}
In the wild images contain richer content and more complex distortions, thus involving more complicated human brain analysis and decision-making mechanisms to assess the image quality, which are usually influenced by human memory. 
Therefore, this paper explores in depth human memory mechanisms during quality rating and models the multi-level features extracted from in the wild images from the perspective of memory. 
Memory involving semantic knowledge, is known as declarative memory \cite{eichenbaum2000cortical}, which is primarily generated in the hippocampus region of the human brain. Depending on the length of memory time, declarative memory can be divided into short-term and long-term memory \cite{cai1990neural,maheu2004differential}. Short-term memory \cite{cowan1988evolving} is formed by temporary storage of attention information, while long-term memory \cite{wood2012review} is formed by repeated reinforcement of short-term memory to form semantic knowledge and make decisions \cite{eichenbaum2000cortical}. 
The human decision about image quality in the wild is not only made based on the information stored in short-term memory, but is also influenced by the experience and knowledge stored in long-term memory.
Therefore, the short-term and long-term memory are simulated via a CNN and a GCN respectively. 

\subsubsection{Short-term Memory Modeling}
Short-term memory can temporarily store attention information and filter redundant information. In this paper, we define a mask $\mathbf M$ to filter the multi-level image features:
\begin{eqnarray}
\mathbf M=\MLP(\mathbf{F}),
\end{eqnarray}
where $\MLP$ is a multi-layer perception (MLP). $\mathbf M\in \mathbb{R}^{C}=\{m_1,m_2,\cdots,m_{C}\}$, in which $C$ is the channel of $\mathbf M$.
$C$ is set to the number of bins of the DOS of image quality in this paper, that is, the number of levels of image quality.
The attention features $\mathbf{AF}$ for $C$ levels of image quality can be written as:
\begin{eqnarray}
\mathbf{AF}=\{\mathrm{AF_1},\mathrm{AF_2},\cdots,\mathrm{AF_C}\}=\ConV(\mathbf{F})^{T} \cdot \mathbf M,
\end{eqnarray}
where $\ConV$ is a convolutional layer with an output channel of $C^{'}$, $\mathbf{AF}\in \mathbb{R}^{C^{'}\times C}$.

After that, the short-term memory mechanism estimates the global semantics $\mathbf{GS}\in \mathbb{R}^{C^{'}\times C}$ of the attention features $\mathbf{AF}$:
\begin{eqnarray}
\mathbf{GS}=\ExP(\ReLU(\GAP(\mathbf{AF}))),
\end{eqnarray}
where $\ReLU$ is a rectified linear unit, and $\ExP$ expands the shape of the input to $C^{'}\times C$. 
Finally, the short-term memory $\mathbf S$ can be written as the concatenation and convolution of the attention features $\mathbf{AF}$ and the global semantics $\mathbf{GS}$:
\begin{eqnarray}
\mathbf S=\{\mathit{S_1},\mathit{S_2},\cdots,\mathit{S_C}\}=\ConV(\mathbf{AF}\oplus \mathbf{GS}).
\end{eqnarray}
Here, the output channel of the convolutional layer $\ConV$ is also set to $C^{'}$. 
The outputs of the short-term memory block are used as inputs to the long-term memory block. The attention features $\mathbf{AF}$ and the short-term memory $\mathbf{S}$ are sent to the gate unit of the long-term memory block to generate the long-term memory of the human brain \cite{yao2022m}.

\subsubsection{Long-term Memory Modeling}
Long-term memory is a persistent memory formed after repeated training of short-term memory, which is influenced by the gating effect of N-Methyl-D-aspartate (NMDA) receptors \cite{morris2013nmda}. Specifically, the outputs of the short-term memory block, i.e. attention features and short-term memory, adjust the excitation threshold in the gate unit to control the formation of long-term memory. 
First, to facilitate the modeling of semantic knowledge (i.e. image quality labels) relationships, it is assumed that the long-term memory shares semantic relationships through a gated GCN \cite{li2015gated}, which uses the vertices in the graph $\mathbf V=[ \mathit{v_1}, \mathit{v_2},\cdots,\mathit{v_C}]$ to represent the $C$ labels of image quality. The adjacency matrix $\mathbf A$ and the weight matrix $\mathbf W$ of the graph together reflect the relationship memory between image quality labels. 

Given the complex nature of brain functions, it is difficult to accurately predict the excitation threshold in the gate unit.
Gating of the excitation threshold in the human brain is facilitated by NMDA receptors, which synchronously monitor the activity of presynaptic and postsynaptic. In light of this, we present a simulation of the gating detection parameter $\mathbf G$ that evaluates the interaction between the input and the current long-term memory. This interaction has the potential to alter the excitation threshold \cite{kong2023multi}. 
Therefore, we propose a modulation of the excitation threshold by changing the input:
\begin{align}
&\mathbf A^{t-1} = f^{a}(\mathbf {AF}^t), ~\mathbf W^{t-1} = f^{w}(\mathbf {S}^t),\\
&\mathbf G_a = \tanh(\mathbf L_a \cdot [\mathbf A^{t-1}, \mathbf {AF}^t]), \\
&\mathbf G_w = \tanh(\mathbf L_w \cdot [\mathbf W^{t-1}, \mathbf {S}^t]),
\end{align}
where $\mathbf A^{t-1}$ and $\mathbf W^{t-1}$ represent the adjacency and weight related to the long-term memory at the previous time point $t-1$, respectively. The functions $f^{a}$ and $f^{w}$ are nonlinear mapping functions. 
$\mathbf {AF}^t$ and $\mathbf {S}^t$ denote the attention features and short-term memory associated with the new input at time point $t$. $\mathbf L_a$ and $\mathbf L_w$ are the weights of the learned linear layers. $\tanh$ is a hyperbolic tangent function. 
$\mathbf G_a$ and $\mathbf G_w$ indicate whether $\mathbf {AF}^t$ and $\mathbf {S}^t$ are activated, whose symbols represent the correlation and the values represent the difficulty of change.

The new degree of stimulation following the influence of relational memory is denoted as:
\begin{align}
&\mathbf Y_{AF} = \mathbf {AF}^t + \mathbf G_a \cdot \mathbf A^{t-1},\\
&\mathbf Y_{S} = \mathbf {S}^t + \mathbf G_w \cdot \mathbf W^{t-1}.
\end{align}
Then, the new adjacency matrix $\mathbf A^{t}$ and weight matrix $\mathbf W^{t}$ are generated in the long-term memory:
\begin{align}
\mathbf A^t = f^{a}(\mathbf Y_{AF}^t), \quad \mathbf W^t = f^{w}(\mathbf Y_{S}^t).
\end{align}
Finally the long-term memory $\mathbf L\in \mathbb{R}^{C^{'}\times C}$ can be written as:
\begin{align}
\mathbf L = \Sigmoid (\mathbf A^t \cdot (\mathbf {AF}^t + \mathbf {S}^t) \cdot \mathbf W^t).
\end{align}

\subsection{Quality Prediction}
In the quality prediction module, we specifically analyze and simulate how human memory is involved in decision-making, which mainly occurs in the hippocampus \cite{palombo2015does,liu2004role}. Specifically, once the image information enters the hippocampus, it reaches the output neurons through two pathways: a direct pathway and an indirect pathway \cite{thierry2000hippocampo,zola1986human}. In the direct pathway, the input information is directly connected to the output neurons. In the indirect pathway, information updates long-term memory by using information from the direct pathway to form semantic knowledge, which then feeds back to the direct pathway for decision-making \cite{yao2022m}. 
The dual-pathway structure facilitates effective information exchange between long-term memory and attention features for decision-making. 
Through the dual-pathway, the memory features can be written as:
\begin{align}
\mathbf M = \mathbf{AF}+\mathbf L, ~\mathbf M\in \mathbb{R}^{C^{'}\times C}.
\end{align}
The predicted memory DOS of image quality is defined as:
\begin{align}
\mathbf{{d}}_{mem}= \Softmax(\GAP(\mathbf M)), ~\mathbf{{d}}_{mem}\in \mathbb{R}^{C}.
\end{align}

In addition to the memory DOS influenced by the SLM mechanism of the human brain, we also consider the DOS predicted completely based on image features extracted by the multi-level feature extraction module:
\begin{align}
\mathbf{d}_{alg} = \Softmax(\FCfirst(\FCsecond(\mathbf{F}))),~\mathbf{d}_{alg}\in \mathbb{R}^{C}, 
\end{align}
where $\FCfirst$ and $\FCsecond$ are full connection layers, and $\mathbf{d}_{alg}$ is called the predicted algorithmic DOS. 
This process, which excludes the influence of subjective memory, is called algorithmic decision-making.


The final predicted DOS of image quality is defined as:
\begin{align}
\mathbf{d}_p=\lambda \cdot \mathbf{d}_{mem} +(1-\lambda) \cdot \mathbf{d}_{alg},
\end{align}
where $\mathbf{d}_p=\{d_{p_1},d_{p_2},\cdots,d_{p_C}\}\in \mathbb{R}^{C}$, $\lambda$ is the hyper-parameter that controls the decision weights between the predicted memory DOS and algorithmic DOS. 
Finally with the predicted DOS, the predicted MOS can be calculated as:
 \begin{eqnarray}
 \label{mos}
\mathrm{MOS_p}=\sum_{c=1}^{C}s_{c}\cdot d_{p_c},
 \end{eqnarray}
where $s_c$ represents the score assigned to the $c$-th image quality label.
In addition, we can also predict the SOS of image quality from the predicted DOS:
 \begin{eqnarray}
\mathrm{SOS_p}=\sqrt{\sum_{c=1}^{C}d_{p_c}\cdot(s_{c}-\mathrm{MOS_p})^2}.
 \end{eqnarray}
\subsection{Multi-label Training Strategy}
\label{Loss Function}
\begin{figure}[t]
	\centering
	\includegraphics[width=3.5in]{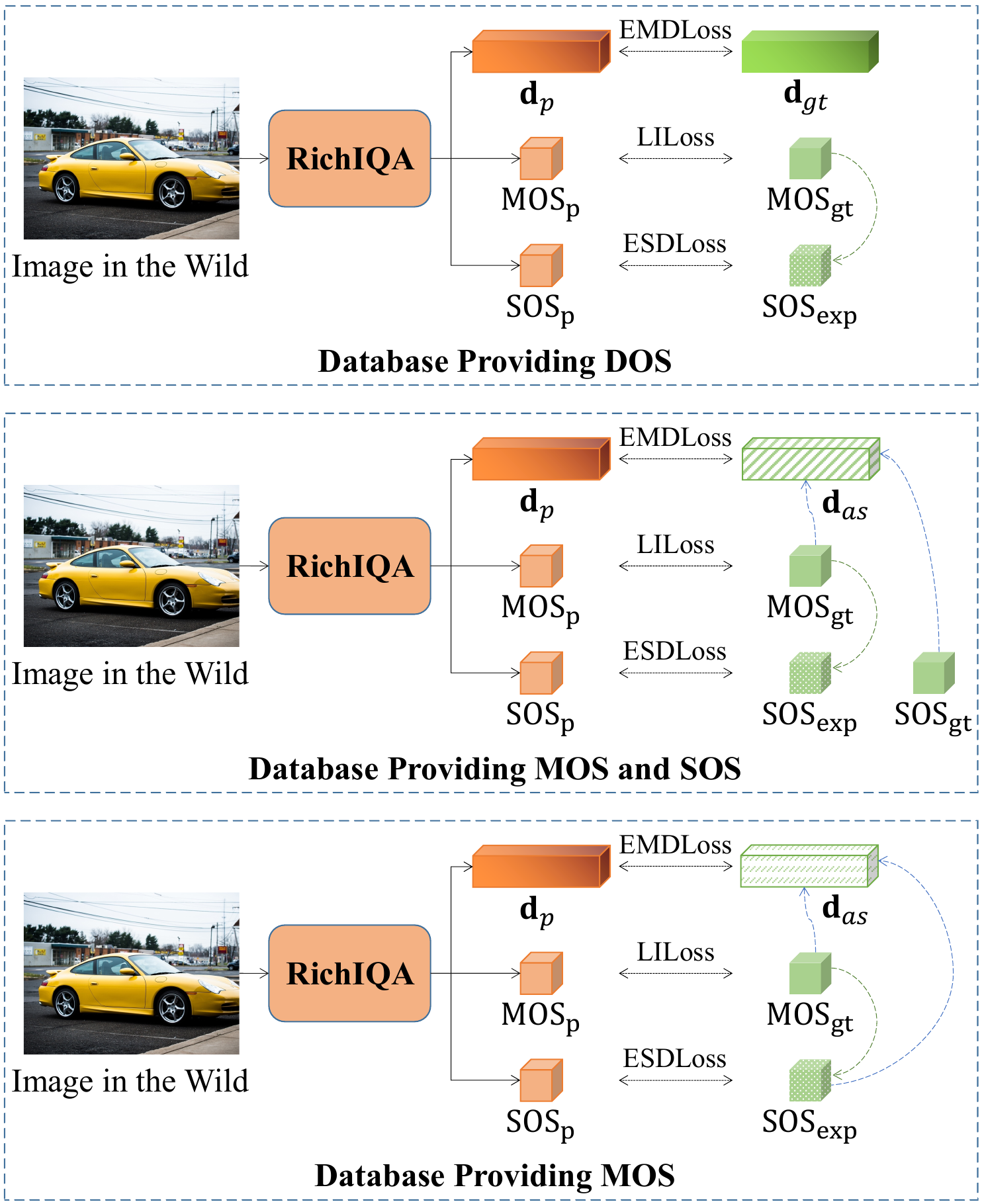}
	\caption{The proposed multi-label training strategy. For databases providing different subjective quality information, we use different subjective quality rating labels and loss functions.}
	\label{fig_loss}
  \vspace{-10pt}
\end{figure}

To train the three-stage network which is tailored to learn the rich subjective rating information, we propose a multi-label training strategy which use the MOS, DOS and SOS of image quality as training labels.
An illustration of the proposed multi-label training strategy is given in Fig. \ref{fig_loss}, where we  use different training labels for databases providing different subjective quality information.

\subsubsection{DOS as Training Label}
 For IQA databases that provide the ground-truth DOS of image quality, we use the earth mover’s distance (EMD) loss function to calculate the distance between the predicted DOS and the ground-truth DOS:
 \begin{eqnarray}
 \label{EMD}
\mathrm{EMDLoss}= \sqrt{\frac{1}{C} \sum_{c=1}^{C}|\sum_{i=1}^{c}{d}_{gt_i}
-
\sum_{i=1}^{c} d_{p_i}|^{2} },
 \end{eqnarray}
where $\mathbf{d}_{gt}=\{d_{{gt}_1},d_{{gt}_2},\cdots,d_{{gt}_C}\}$ is the ground-truth DOS of image quality provided by IQA databases. 
It is important to note that some IQA databases may not provide the ground-truth DOS of image quality.
To improve the applicability of our proposed method for IQA databases, we use different methods to supplement the missing DOS of image quality.

For IQA databases that provide the ground-truth MOS and SOS of image quality, we first assume that the DOS of image quality follows a Gaussian distribution:
 \begin{eqnarray}
 \label{gd1}
d_{as_c}=\mathcal{N}(s_c;\mathrm{MOS_{gt}},\mathrm{SOS_{gt}^2}),~\Sigma_{c=1}^C d_{as_c}=1,
 \end{eqnarray}
 where $\mathbf{d}_{as}=\{d_{{as}_1},d_{{as}_2},\cdots,d_{{as}_C}\}$ is the assumed DOS of image quality, $\mathrm{MOS_{gt}}$ is the ground-truth MOS of image quality, and $\mathrm{SOS_{gt}}$ is the ground-truth SOS of image quality.
 Then, we use the EMD loss function in Eq. (\ref{EMD}) to calculate the distance between the predicted DOS of image quality $\mathbf{d}_{p}$ and the assumed DOS of image quality  $\mathbf{d}_{as}$.

  For IQA databases that only provide the ground-truth MOS of image quality, we first estimate the  expected SOS of image quality according to \cite{2011SOS}:
 \begin{align} 
 \label{calsos}
\mathrm{SOS_{exp}^2} =&\; a\cdot(-\mathrm{MOS_{gt}^2}+(\mathrm{S_{start}}+\mathrm{S_{end}})\cdot\mathrm{MOS_{gt}} \notag \\
&\; -\mathrm{S_{start}}\cdot \mathrm{S_{end}}),
 \end{align}
where $a$ is an empirical parameter, $\mathrm{S_{start}}$ and $\mathrm{S_{end}}$ denote that the score range of image quality scores in the IQA database goes from $\mathrm{S_{start}}$ to $\mathrm{S_{end}}$.
Then, we assume that the DOS of image quality follows the Gaussian distribution:
   \begin{eqnarray}
 \label{gd2}
d_{as_c}=\mathcal{N}(s_c;\mathrm{MOS_{gt}},\mathrm{SOS_{exp}^2}),~\Sigma_{c=1}^C d_{as_c}=1.
 \end{eqnarray}
Similarly, we can still use the EMD loss function in Eq. (\ref{EMD}) to calculate the distance between the predicted DOS of image quality $\mathbf{d}_{p}$ and the assumed DOS of image quality $\mathbf{d}_{as}$.

\subsubsection{MOS as Training Label}
Considering that our method can also predict the MOS and SOS of image quality, we first use a L1 loss function to calculate the distance between the predicted MOS and the ground-truth MOS to ensure that the predicted image quality is consistent with the perceptual image quality:
 \begin{eqnarray}
  \label{L1}
\mathrm{L1Loss}= |\mathrm{MOS_{gt}}-\mathrm{MOS_{p}}|.
 \end{eqnarray}

Obviously, when the image quality is relatively high or low, subjects may show less diversity when evaluating the image, indicating that the SOS of image quality is relatively small. Conversely, when the image quality is in the middle, subjects may display a greater diversity when evaluating the image, which means that the SOS of image quality is relatively large. This is consistent with the relationship between MOS and SOS shown in Eq. (\ref{calsos}). 
To ensure that our method can successfully learn subjective diversity,
we design an expected standard deviation (ESD) loss function to calculate the distance between the predicted SOS and the expected SOS:
\begin{eqnarray}
\label{sos}
\mathrm{ESDLoss}= |\mathrm{SOS_{exp}}-\mathrm{SOS_{p}}|^2.
\end{eqnarray}




\subsubsection{Total Loss}
All three loss functions described above (Eq. (\ref{EMD}), Eq. (\ref{L1}), and Eq. (\ref{sos})) are combined for joint training of the proposed network:
  \begin{eqnarray}
\mathrm{Loss}= \alpha \cdot\mathrm{EMDLoss}+\beta \cdot \mathrm{L1Loss}+ \gamma \cdot\mathrm{ESDLoss},
 \end{eqnarray}
 where $\alpha$, $\beta$, and $\gamma$ represent the balance factors of the three loss functions, respectively.

\section{Experiments}
\label{sec:result}
In this section, we first introduce the used in the wild IQA databases and the experimental protocol, then report the comparison results of RichIQA with other state-of-the-art IQA methods, and finally demonstrate the effectiveness and feasibility of RichIQA via extensive analysis experiments.

\subsection{In the Wild IQA Databases}
\label{database}
A total of five mainstream in the wild IQA databases are used for experimental validation. An overview of the five databases is given in Table \ref{table_database}. 
\subsubsection{KonIQ-10K}
The KonIQ-10K \cite{2019KonIQ} database is an in the wild IQA database which contains 10,073 images collected from publicly available resources on the Internet to ensure its diversity and practicality. These images contain various authentic distortions that are not specifically designed but occur naturally. The quality score of each image is obtained through crowdsourcing, with thousands of observers rating the image based on its visual quality. 
In the released database, both the MOS ranging from 1 to 5 and the DOS of image quality are provided. 
The KonIQ-10K database can effectively reflect the breadth and complexity of image quality changes in the real world, improving the practicality and generalizability of IQA methods.
\subsubsection{BID}
The BID \cite{ciancio2010no} database is specifically designed for blurred images, which is an important research direction in the field of IQA. The BID database contains 6,000 images covering a wide range of scenes, including natural landscapes, urban landscapes, indoor scenes, \textit{etc.}, in order to simulate the different conditions and contexts that may be encountered in the real world.
The database includes a variety of blur types, including motion blur, lens blur, and others. It also includes a spectrum of blur levels, ranging from mild blur to extreme blur.
To assess the degree of image blur, each image in the BID is annotated in detail, on a scale of 0 to 5. 
The publicly available database contains the scores assigned to each image by each observer, which allows us to obtain the MOS and DOS of image quality.
\subsubsection{LIVE Challenge}
The LIVE Challenge \cite{ghadiyaram2015massive} database is a large-scale database used for in the wild IQA research, with a particular focus on the use of crowd-sourcing to collect subjective quality scores. This database is widely used in the image and video processing community to test and validate various IQA methods. The LIVE Challenge database contains 1,162 natural images exhibiting a wide range of authentic distortions that occur during the capture and storage in the natural environment.  The quality score of each image is crowd-sourced, with thousands of different observers rating the visual quality of these images on online platforms with a scale of 0 to 100. The authors have made the MOS and SOS of each image publicly available.
\subsubsection{SPAQ}
The SPAQ \cite{fang2020perceptual} database
is specifically designed to evaluate the quality of images captured by smartphones. The database contains 11,125 images taken with smartphones. These images cover a variety of scenes and conditions, including outdoor natural scenes, indoor portraits, and other subjects. The distortions of the images in this database are attributed to limitations inherent to smartphones, such as sensor quality, image processing algorithms, and other factors. In the SPAQ database, observers are asked to rate the perceptual quality of images on a scale from 0 to 100, which reflects the actual user perception of smartphone imaging quality. The authors release only the MOS of image quality.
\subsubsection{FLIVE}
The FLIVE \cite{ying2020patches} database is a database used for in the wild IQA research. Its objective is to map the perceptual space of image quality by extending the perceptual quality assessment of local image blocks (patches) to the entire image. The database contains 39,810 authentically distorted images from a variety of sources, including natural landscapes, urban landscapes, portraits, daily life, and other categories, to ensure wide scene coverage. The quality score of each image is also obtained through crowd-sourcing. Observers are required to evaluate the visual quality of these images on a scale from 0 to 100. The authors release only the MOS of image quality.

  \begin{table}[t]
	\renewcommand\arraystretch{1.3}
	\tabcolsep=0.32cm
	\begin{center}
		\caption{An Overview of the Five in the Wild IQA Databases.
		} \label{table_database}
		\begin{tabular}{c||c|c|c}
			\toprule[0.8pt]
			Database & Image \# & Range & Label\\ 
			\noalign{\vskip 0.5ex}\toprule[0.8pt]
			KonIQ-10K \cite{2019KonIQ} & 10,073 & [1,5] & MOS, DOS\\  
			BID \cite{ciancio2010no} & 6,000 & [0,5] & MOS, DOS\\
			LIVE Challenge \cite{ghadiyaram2015massive}  & 1,162 & [0,100] & MOS, SOS\\
			SPAQ \cite{fang2020perceptual} & 11,125 & [0,100] & MOS\\
			FLIVE \cite{ying2020patches} & 39,810 & [0,100] & MOS\\
			\bottomrule[0.8pt]
		\end{tabular}
	\end{center}
 \vspace{-10pt}
\end{table}

   \begin{table*}[t]
\renewcommand\arraystretch{1.3}
\tabcolsep=0.55cm
\begin{center}
\caption{Performance Comparison of Predicting the \textbf{MOS} of Image Quality on the \textbf{KonIQ-10K} and \textbf{BID} Databases. The Best and Second-best Performances are in Bold and Underlined Respectively. Similar Formatting Rules Apply to the Following Tables.
} \label{t1}
\begin{tabular}{c||c| c|c|c|c|c}
  \toprule[0.8pt]
        &\multicolumn{3}{c|}{KonIQ-10K}&\multicolumn{3}{c}{BID}\\
        \noalign{\vskip 0.5ex}\toprule[0.5pt]
        Method & SRCC $\uparrow$ & PLCC $\uparrow$ &  RMSE $\downarrow$ & SRCC $\uparrow$ & PLCC $\uparrow$ &  RMSE $\downarrow$\\
        \noalign{\vskip 0.5ex}\toprule[0.8pt]
        BRISQUE \cite{saad2012blind} & 0.6954 & 0.6903 & 0.4037 &0.4327  &  0.4492   & 1.1243\\  
        NIQE \cite{mittal2012making}  & 0.6692 & 0.6624 & 0.4178& 0.3891  &  0.3801  &  1.4324 \\  
        BMPRI \cite{min2018blind}& 0.4269 & 0.4246 & 0.5110 & 0.3630 &   0.3604   & 1.1836\\  
        CNNIQA \cite{kang2014convolutional}& 0.5711 & 0.5071 & 0.4993 &0.6001&0.5715&1.0842\\  
        DIQaM-NR \cite{bosse2017deep} & 0.7901 & 0.7734 & 0.4201&0.5494&0.5621&1.1012 \\  
        WaDIQaM-NR \cite{bosse2017deep} & 0.8484 & 0.8452 & 0.3272&0.4519&0.4418&1.1442 \\    
        DBCNN \cite{zhang2018blind} & 0.8594 & 0.8787 & 0.2743 &0.7737&0.7806&0.8223\\  
        HyperIQA \cite{su2020blindly}  & 0.9040 & 0.9163 & 0.2250 &\underline{0.8779}&\underline{0.8826}&0.6103\\ 
        TReS \cite{golestaneh2021no}& \underline{0.9100}&\underline{0.9220}&\underline{0.2006}&0.7556&0.7462& 1.1728\\
        DACNN \cite{pan2022dacnn}  & 0.8956 & 0.9121 & 0.2309 &0.7635&0.7685&0.8259\\  
        GraphIQA \cite{sun2022graphiqa} &           0.8389 & 0.8609 & 0.3548 &0.7846&0.7903&0.7914\\  
         NIMA \cite{2017NIMA}& 0.7803 & 0.7868 & 0.3428 &0.6610&0.6823&0.8530\\  
        Gao \cite{gao2022image}& 0.9045 & 0.9185 & 0.2185& 0.8503&0.8789&\underline{0.5462}\\ 
        StairIQA \cite{sun2023blind}&0.9083&0.9167&0.2070&0.8306&0.8624&0.6367\\
        REQA \cite{li2024reqa} &0.8225&0.8447&0.3696&0.7408&0.7493&0.9237 \\
 \noalign{\vskip 0.5ex}\cdashline{1-7}[2pt/2pt]\noalign{\vskip 1.0ex}
        \textbf{RichIQA} &\textbf{0.9383}&\textbf{0.9500}& \textbf{0.1727}&  \textbf{0.8998}&\textbf{0.9085}&\textbf{0.5159}\\
\bottomrule[0.8pt]
\end{tabular}
\end{center}
 \vspace{-10pt}
\end{table*}

\subsection{Experimental Protocol}
\subsubsection{Implementation Detail}
The experiments are performed on the five in the wild IQA databases described above. During the experiment, 80$\%$ of the images in the database are divided into the training set, and the remaining 20$\%$ of the images are divided into the test set. 
The database is randomly split for 10 times, and the mean value of the results is reported.
We first resize the resolution of the images to $512\times 512$ and then randomly crop images with a resolution of $384\times 384$. The proposed method is implemented in PyTorch. The Adam optimizer \cite{2014Adam} with an initial learning rate of 0.00001 is used to train the proposed method on a server with NVIDIA GTX 4090. The batch size is set to 8.

The weights of the CvT are initialized by training on ImageNet \cite{deng2009imagenet}. In this paper, $C_1=64$, $H_1=96$, and $W_1=96$. $C_2=192$, $H_2=48$, and $W_2=48$. $C_3=384$, $H_3=24$, and $W_3=24$. $C$ is set to 5, and $C^{'}$ is set to 256. $\lambda=0.999$, and $\alpha:\beta:\gamma=200:10:1$. 
In addition, by substituting the $\mathrm{SOS_{exp}}$ in Eq. (\ref{calsos}) with the ground-truth SOS provided by the database, we calculate the $a$ for the KonIQ-10K, BID, and LIVE Challenge databases.
Specifically, the $a$ for KonIQ-10K is 0.0907; the $a$ for BID is 0.1683; and the $a$ for LIVE Challenge is 0.1841; while the average value of $a$ for these three databases is 0.1477. Therefore, we set $a$ to 0.1477 in this paper.

  \begin{table}[thp]
\renewcommand\arraystretch{1.3}
\tabcolsep=0.45cm
\begin{center}
\caption{Performance Comparison of Predicting the \textbf{MOS} of Image Quality on the \textbf{LIVE Challenge} Database.
} \label{t2}
\begin{tabular}{c||c|c|c}
  \toprule[0.8pt]
        &\multicolumn{3}{c}{LIVE Challenge}\\ 
        \noalign{\vskip 0.5ex}\toprule[0.5pt]
        Method & SRCC$\uparrow$ & PLCC$\uparrow$ & RMSE$\downarrow$\\ 
        \noalign{\vskip 0.5ex}\toprule[0.8pt]
        BRISQUE \cite{saad2012blind}  & 0.5814 & 0.6039&16.1273\\  
        NIQE \cite{mittal2012making} & 0.5803 & 0.5987&16.3556 \\  
        BMPRI \cite{min2018blind}&0.3724 & 0.3919 &18.9930 \\  
        CNNIQA \cite{kang2014convolutional}& 0.6612 & 0.6239&15.4325 \\  
        DIQaM-NR \cite{bosse2017deep} & 0.5950 & 0.5936 &15.3241 \\  
        WaDIQaM-NR \cite{bosse2017deep}  & 0.6544 & 0.6597&15.0178 \\   
        DBCNN \cite{zhang2018blind}& 0.8314 & 0.8545 &10.6661\\  
        HyperIQA \cite{su2020blindly}& 0.8453 & 0.8589&11.2274 \\ 
        TReS \cite{golestaneh2021no}&\underline{0.8880}&0.8523&\underline{8.1762}\\
        DACNN \cite{pan2022dacnn} & 0.8485 & 0.8602 & 10.4241\\ 
        GraphIQA \cite{sun2022graphiqa}& 0.8079 & 0.8335& 12.4325\\  
         NIMA \cite{2017NIMA} & 0.7811 & 0.8124&13.8946 \\  
        Gao \cite{gao2022image} & 0.7706 & 0.8012 &13.3523  \\ 
        StairIQA \cite{sun2023blind}&0.8460& \underline{0.8685}&{9.7764}\\
        REQA  \cite{li2024reqa} &{0.8166}&{0.8331}&{11.6510} \\
  \noalign{\vskip 0.5ex}\cdashline{1-4}[2pt/2pt]   \noalign{\vskip 1.0ex}
        \textbf{RichIQA} &\textbf{0.8943}&\textbf{0.9121}& \textbf{8.2312}\\
\bottomrule[0.8pt]
\end{tabular}
\end{center}
 \vspace{-10pt}
\end{table}

    \begin{table*}[t]
\renewcommand\arraystretch{1.3}
\tabcolsep=0.55cm
\begin{center}
\caption{Performance Comparison of Predicting the \textbf{MOS} of Image Quality on the \textbf{SPAQ} and \textbf{FLIVE} Databases.
} \label{t3}
\begin{tabular}{c||c|c|c|c|c|c}
  \toprule[0.8pt]
        &\multicolumn{3}{c|}{SPAQ}&\multicolumn{3}{c}{FLIVE}\\ 
       \noalign{\vskip 0.5ex} \toprule[0.5pt]
        Method & SRCC$\uparrow$  & PLCC$\uparrow$&RMSE$\downarrow$ & SRCC$\uparrow$  & PLCC$\uparrow$ &RMSE$\downarrow$\\ 
        \noalign{\vskip 0.5ex} \toprule[0.8pt]
        BRISQUE  \cite{saad2012blind}& 0.7075 & 0.7080 & 14.7846 &0.1864  &  0.2150  & 6.0886\\ 
        NIQE  \cite{mittal2012making} & 0.7002 & 0.7010 &14.9751& 0.1838 &   0.2066  &  6.0589 \\ 
        BMPRI \cite{mittal2012making}& 0.6083 & 0.6110 &16.7090 &0.1900  &  0.2137  &  6.0728\\  
        CNNIQA \cite{kang2014convolutional}&0.6834& 0.6848&14.1684&0.2924 &0.3284&6.0949\\
        DIQaM-NR \cite{bosse2017deep}&0.8240&0.8360&11.1463&0.4002&0.3929 &6.1745\\
        WaDIQaM-NR \cite{bosse2017deep} &0.8210&0.8430&12.2562&0.4553&0.4768 &5.6717\\
        DBCNN  \cite{zhang2018blind}& 0.8847 & 0.8892 &9.7329 &0.3676&0.4542&5.6351\\  
        HyperIQA \cite{zhang2018blind}& 0.7528 & 0.7564 &16.5553& 0.5308&0.6193&4.7990 \\  
        TReS \cite{golestaneh2021no}&0.8438&0.8440& 10.4234&0.6000&0.5525&11.8705 \\
        DACNN \cite{pan2022dacnn} & 0.8870 & 0.8912 &8.2973&0.5502&0.6307&4.8224 \\  
        GraphIQA \cite{sun2022graphiqa}& 0.7690 & 0.7670 &11.3524& 0.4469&0.5402&5.3329\\  
        NIMA \cite{2017NIMA}& 0.7621&	0.7666& 12.3741&0.5198&0.6110&5.2254\\  
        Gao \cite{gao2022image}& \underline{0.9099} & \underline{0.9131}&8.5313&0.5556&0.6369&\underline{4.5822}\\ 
        StairIQA \cite{sun2023blind}&0.9078 & 0.9113&\underline{8.0579}&\underline{0.5563}&\underline{0.6435}&4.6040\\
        REQA  \cite{li2024reqa} &0.7541& 0.7566 &13.8490&0.5065&0.5703&6.6243\\
\noalign{\vskip 0.5ex} \cdashline{1-7}[2pt/2pt] \noalign{\vskip 1.0ex}
        \textbf{RichIQA} &\textbf{0.9232}&\textbf{0.9268}& \textbf{7.8559}&\textbf{0.5828}&   \textbf{0.6843}&     \textbf{4.4867}\\
\bottomrule[0.8pt]
\end{tabular}

\end{center}
 \vspace{-10pt}
\end{table*}

    \begin{table*}[thp]
\renewcommand\arraystretch{1.3}
\tabcolsep=0.3cm
\begin{center}
\caption{Performance Comparison of Predicting the \textbf{DOS} of Image Quality on the \textbf{KonIQ-10K} and \textbf{BID} Databases.
} \label{distribution}
\begin{tabular}{c||c|c|c|c|c|c|c|c|c|c}
  \toprule[0.8pt]
  &\multicolumn{5}{c|}{KonIQ-10K}&\multicolumn{5}{c}{BID}\\
  \noalign{\vskip 0.5ex}\hline\noalign{\vskip 0.5ex}
  Method &JSD$\downarrow$ & EMD$\downarrow$ & RMSE$\downarrow$ & Intersect.$\uparrow$ &Cosine$\uparrow$    &JSD$\downarrow$ & EMD$\downarrow$ & RMSE$\downarrow$ & Intersect.$\uparrow$ &Cosine$\uparrow$\\
  \noalign{\vskip 0.5ex} \toprule[0.8pt]
  NIMA   
&0.0415&0.0835&0.1087&0.8073&0.9267&0.2203&0.2831&0.2461&0.4981&0.5971\\
  Liu  
  &0.0214&0.0952& 0.1045&0.7950&0.9167&0.1589&0.2263&0.1917&0.5957&0.7597\\
KonCept512&          0.0624&0.1074&0.1214&0.7747&0.9032&0.1025&0.1505&0.1547&0.6917&0.8306\\
Gao& \underline{0.0213} & \underline{0.0604} & \underline{0.0719} & \underline{0.8687} & \underline{0.9622} &\underline{0.0796}&\underline{0.1290}&\underline{0.1325}&\underline{0.7361}&\underline{0.8733}\\
 \noalign{\vskip 0.5ex} \cdashline{1-11}[2pt/2pt] \noalign{\vskip 1.0ex}
\textbf{RichIQA} & \textbf{0.0149}& \textbf{0.0494}&\textbf{0.0599}&\textbf{0.8914}&\textbf{0.9740}&\textbf{0.0758}&\textbf{0.1192}&\textbf{0.1266}&\textbf{0.7450}&\textbf{0.8864}\\ 
\bottomrule[0.8pt]
\end{tabular}
\end{center}
 \vspace{-10pt}
\end{table*}

\subsubsection{Competing Methods and Evaluation Criteria}
When predicting the MOS of image quality,  
we use BRISQUE \cite{saad2012blind}, NIQE \cite{mittal2012making}, BMPRI \cite{min2018blind}, CNNIQA \cite{kang2014convolutional}, (Wa)DIQaM-N \cite{bosse2017deep},  DBCNN \cite{zhang2018blind}, HyperIQA \cite{su2020blindly}, TReS \cite{golestaneh2021no}, DACNN \cite{pan2022dacnn}, GraphIQA \cite{sun2022graphiqa}, NIMA \cite{2017NIMA}, Gao \cite{gao2022image}, StairIQA \cite{sun2023blind}, and REQA  \cite{li2024reqa} as competing methods.
The Spearman rank order correlation coefficient (SRCC), Pearson linear correlation coefficient (PLCC), and root mean square error (RMSE) are used as evaluation criteria.
Before calculating the PLCC, we use a four parameter logistic function to the predicted MOS:
\begin{equation}\label{four-parameter}
\mathrm{MOS_m}=\frac{\beta_1-\beta_2}{1+\exp{(-(\mathrm{MOS_p}-\beta_3)/\beta_4)}}+\beta_2,
\end{equation}
where $\mathrm{MOS_m}$ is the mapped $\mathrm{MOS_p}$, and $\{ \beta_i|i=1,2,3,4\}$ are parameters fitted with the least square loss between $\mathrm{MOS_m}$ and $\mathrm{MOS_p}$.
The closer the SRCC and PLCC are to 1, the better the prediction performance. The closer the value of RMSE is to 0, the better the prediction performance.

When predicting the DOS of image quality, the following methods are used as competing methods:
NIMA \cite{2017NIMA}, Liu \cite{2018Comprehensive},  KonCept512 \cite{2019KonIQ}, and Gao \cite{gao2022image}. 
The Jensen-Shannon distance
(JSD), EMD, RMSE, intersection, and cosine are used as evaluation criteria \cite{gao2022image}. 
The closer the values of intersection and cosine are to 1, the better the prediction performance of the method. The closer the values of JSD, EMD, and RMSE are to 0, the better the prediction performance of the method.

    \begin{table*}[t]
	\renewcommand\arraystretch{1.32}
	\tabcolsep=0.40cm
	\begin{center}
		\caption{ Performance Comparison of Predicting the \textbf{MOS} of Image Quality When \textbf{Trained on the KonIQ-10K and BID Databases} and \textbf{Tested on Other Databases}.
		} \label{Cross-Database_kon}
		\begin{tabular}{c||c|c|c|c}
			\toprule[0.8pt]
			Training Database (Criteria)&\multicolumn{4}{c}{KonIQ-10K (SRCC$\uparrow$ / PLCC$\uparrow$)}\\
			\noalign{\vskip 0.5ex} \toprule[0.5pt]
			Test Database  &BID&LIVE Challenge&  SPAQ &FLIVE\\ 
			\noalign{\vskip 0.5ex} \toprule[0.8pt]
			BRISQUE  \cite{saad2012blind}& 0.4582 /  0.4553&0.3153 / 0.3288 	&	 0.3393 / 0.3356& 0.2432 / 0.3051\\ 
			NIQE  \cite{mittal2012making} & 0.4475 /     0.4422&0.5071 / 0.5397&	 0.3063 / 0.2935& 0.2354 / 0.2964 \\ 
			BMPRI \cite{mittal2012making}& 0.2028 / 0.1886&0.3193 / 0.3452& 0.3981 / 0.3694& 0.0604 / 0.0846\\  
			DBCNN  \cite{zhang2018blind}& 0.8051 / 0.8005 &	0.7450 / 0.7797 	&	0.8098 / 0.8090 & 	0.4390 / 0.4986\\  
			HyperIQA \cite{zhang2018blind}& \underline{0.8107} / 0.8051 &0.7707 / 0.7954& 0.2221 / 0.2369&0.4460 / 0.5040\\  
			TReS \cite{golestaneh2021no}& 0.5989 / 0.5772 &0.7998 
			/ 0.7770&0.3469 / 0.3514&0.4157 / 0.4483\\
			DACNN \cite{pan2022dacnn} &0.6158 / 0.5942&0.7789 / 0.7951&\underline{0.8182} / 0.8418&\underline{0.4464} / \underline{0.5172} \\  
			GraphIQA \cite{sun2022graphiqa}& 0.6452 / 0.6359 &	0.7309 / 0.7516  	& 0.1606 / 0.2986&  0.4197 / 0.4806\\  
			Gao \cite{gao2022image}&0.8098 / \underline{0.8124}&0.7999 / 0.7700& {0.8144} / 0.8544 & 0.4411 / 0.4954\\ 
			StairIQA \cite{sun2023blind}&0.8017 / 0.8037 &\underline{0.8018} / \underline{0.8285}&0.8141 / \underline{0.8715}&0.4275 / 0.4966 \\
			REQA  \cite{li2024reqa} & 0.6807 / 0.6687	&0.6948 / 0.7095 &0.2388 / 0.2292&0.3837 / 0.4385 \\
			\noalign{\vskip 0.5ex}\cdashline{1-5}[2pt/2pt]\noalign{\vskip 1.0ex}
			\textbf{RichIQA} & \textbf{0.8711} /  \textbf{0.8793}&\textbf{0.8411} / \textbf{0.8718}&\textbf{0.8769} / \textbf{0.8796}&\textbf{0.4657} / \textbf{0.5403}\\ 
			\bottomrule[0.8pt]\noalign{\vskip 1.0ex}
			
			Training Database (Criteria)&\multicolumn{4}{c}{BID (SRCC$\uparrow$ / PLCC$\uparrow$)}\\ 
			\noalign{\vskip 0.5ex}\toprule[0.5pt]
			Test Database  &KonIQ-10K &LIVE Challenge&  SPAQ &FLIVE\\ 
			\noalign{\vskip 0.5ex}\toprule[0.8pt]
			BRISQUE  \cite{saad2012blind}&0.5009 / 0.5045& 0.2998 / 0.2857 &0.4916 / 0.4907 &0.1404 / 0.1718  \\ 
			NIQE  \cite{mittal2012making} &0.4904 / 0.4955 & 0.3428 /       0.3915&0.4376 / 0.4409& 0.1656 / 0.1993 \\ 
			BMPRI \cite{mittal2012making}& 0.1175 / 0.1281 & 0.1603 /      0.1875 &0.3317 / 0.3285&0.0397 / 0.0312  \\  
			DBCNN  \cite{zhang2018blind}&0.5382 / 0.6086&0.6131 / 0.6659&0.7662 / 0.7687&0.2858 / 0.3815\\  
			HyperIQA \cite{zhang2018blind}& \underline{0.6887} / \underline{0.7368}&\underline{0.7613} / 0.8156&0.1866 / 0.1871&0.3039 / 0.3831\\  
			TReS \cite{golestaneh2021no}& 0.6011 / 0.5189&0.6787 / 0.6140 &0.6563 / 0.6564&0.3078 / 0.4114 \\
			DACNN \cite{pan2022dacnn} & 0.3736 / 0.4347&0.4744 / 0.5422&0.6401 / 0.6437&0.2206 / 0.2923\\  
			GraphIQA \cite{sun2022graphiqa}& 0.5009 / 0.5413&0.5731 / 0.6189&0.2972 / 0.3019&0.2699 / 0.3317\\  
			Gao \cite{gao2022image}& 0.6682 / 0.7251&0.7250 / 0.7622&0.8097 / 0.8223&0.3081 / 0.3923\\ 
			StairIQA \cite{sun2023blind}& 0.6713 / 0.7317&0.7594 / \underline{0.8183}&\underline{0.8338} / \underline{0.8340}&\underline{0.3087} / \underline{0.4237}\\
			REQA  \cite{li2024reqa} & 0.5351 / 0.5585&0.5890 / 0.6255&0.2330 / 0.2389&0.2371 / 0.2829 \\
			\noalign{\vskip 0.5ex}\cdashline{1-5}[2pt/2pt]\noalign{\vskip 1.0ex}
			\textbf{RichIQA} & \textbf{0.7773} / \textbf{0.8201}&\textbf{0.7939} / \textbf{0.8412}&\textbf{0.8343} / \textbf{0.8386}&\textbf{0.3648} / \textbf{0.4722} \\ 
			\bottomrule[0.8pt]
		\end{tabular}
	\end{center}
 \vspace{-10pt}
\end{table*}

\subsection{Comparison Results}
The five in the wild IQA databases used can be divided into three categories based on the provided subjective data: 
\begin{itemize}
\item Databases that provide the DOS of image quality: KonIQ-10K and BID.
\item  Databases that provide the MOS and the SOS of image quality: LIVE Challenge.
\item Databases that provide only the MOS of image quality: SPAQ and FLIVE.
\end{itemize}
The proposed RichIQA method requires the network to be trained jointly using the DOS and MOS. For databases that do not provide the DOS of image quality, we propose to use different methods to supplement the missing DOS, as described in Section \ref{Loss Function}.

\subsubsection{MOS Prediction}
We first evaluate the ability of RichIQA and competitors for predicting the MOS of image quality across the three categories of in the wild IQA databases. The results are given in Tables \ref{t1}, \ref{t2}, and \ref{t3}. 
Table \ref{t1} shows the performances of RichIQA and competing methods in predicting the MOS of image quality on the KonIQ-10K and BID databases. RichIQA uses the DOS and MOS provided by the databases to train the network. As shown in the table, RichIQA significantly outperforms the competing methods, bringing 3.1\%, 3.0\% SRCC gains (compared with the second-best, the same rule applies to the following performance gain numbers) on KonIQ-10K and BID databases respectively. 
Table \ref{t2} presents the performances of RichIQA and competing methods in predicting the MOS of image quality on the LIVE Challenge database. In this case, RichIQA uses the MOS of image quality provided by the database and the Gaussian-based DOS of image quality calculated from Eq. (\ref{gd1}) to train the network. The results show that RichIQA outperforms the competitors, especially in terms of PLCC (with 5.0\% gain). 
Table \ref{t3} lists the performances of RichIQA and the competing methods in predicting the MOS of image quality on the SPAQ and FLIVE databases. Here, RichIQA uses the MOS of image quality provided by the databases and the DOS of image quality calculated from Eq. (\ref{gd2}) to train the network. The results indicate that RichIQA significantly outperforms the competitors in terms of various criteria, with 1.5\%, 4.8\% SRCC gains on SPAQ and FLIVE databases respectively. 
From the experimental results in Tables \ref{t1}, \ref{t2}, and \ref{t3}, it can be concluded that RichIQA demonstrates superior prediction performances in predicting the MOS of image quality, regardless of the subjective data type provided by the IQA databases. This highlights the broad applicability and effectiveness of RichIQA. 

\subsubsection{DOS Prediction}
RichIQA can predict not only the MOS but also the DOS of image quality, which is a byproduct of RichIQA and also a capability rarely owned by the existing mainstream IQA measures. 
We additionally compare the performances of RichIQA and competing methods in predicting the DOS of image quality on KonIQ-10K and BID databases, which are the two only in the wild IQA databases with ground-truth DOS quality labels. The comparison results are given in Table \ref{distribution}, from which it can be observed that the DOSs predicted by RichIQA are closer to human perceptions than other competitors. Compared with the second-best performing method, which is a specifically designed DOS prediction model, 2.6\%, 1.2\% intersection gains are observed on KonIQ-10K and BID databases respectively. 
In summary, compared to the competitors that directly predict the MOS of image quality, RichIQA not only has higher prediction performance, but also can obtain richer subjective information about image quality.


  \begin{table*}[thp]
	\renewcommand\arraystretch{1.3}
	\tabcolsep=0.3cm
	\begin{center}
		\caption{Impact of \textbf{Using Different Stages of CvT} on the Prediction Performance of RichIQA. `\checkmark' Indicates that the Stage is Included. `\ding{55}' Indicates that the Stage is not Included.
		} \label{CvT}
		\begin{tabular}{c|c|c||c|c |c|c |c|c |c|c |c|c}
			\toprule[0.8pt]
			\multicolumn{3}{c||}{ Stage } &\multicolumn{2}{c|}{KonIQ-10K }&\multicolumn{2}{c|}{BID}&\multicolumn{2}{c|}{LIVE Challenge}&\multicolumn{2}{c|}{SPAQ}&\multicolumn{2}{c}{FLIVE}\\ 
			\noalign{\vskip 0.5ex}\toprule[0.5pt]
			1& 2& 3 & SRCC$\uparrow$  & PLCC$\uparrow$  & SRCC$\uparrow$  & PLCC$\uparrow$  & SRCC$\uparrow$  & PLCC$\uparrow$  & SRCC$\uparrow$  & PLCC$\uparrow$  & SRCC$\uparrow$  & PLCC$\uparrow$ \\
			\noalign{\vskip 0.5ex}\toprule[0.8pt]
			\checkmark&\ding{55}&\ding{55}&0.7303&0.7818&0.5487& 0.5520&0.5430&0.5507&0.8492&0.8546&0.4849&0.5077\\
			\ding{55}&\checkmark&\ding{55}&0.9065& 0.9162&0.8325&0.8513&0.7859&0.8011&0.9142& 0.9191&0.5476&0.6302\\
			\ding{55}&\ding{55}&\checkmark&\underline{0.9369}&\underline{0.9480}&\underline{0.8852}&0.8954&0.8883& 0.9023&\underline{0.9230}&0.9233&\underline{0.5813}&\underline{0.6830}\\
			\checkmark&\checkmark&\ding{55}&0.9012&0.9120&0.7730&0.7740&0.7386& 0.7537&0.9154&0.9188& 0.5394& 0.6096 \\
			\checkmark&\ding{55}&\checkmark&0.9353& 0.9472& 0.8797&0.9004&\underline{0.8914}&\underline{0.9073}&0.9225&\underline{0.9259}& 0.5804  &        0.6747\\
			\ding{55}&\checkmark&\checkmark&0.9356& 0.9473&0.8809& \underline{0.9012} &0.8904&0.9051&0.9222& \underline{0.9259}&0.5701&0.6691\\
			
			\noalign{\vskip 0.5ex}\cdashline{1-13}[2pt/2pt]\noalign{\vskip 1.0ex}
			
			\checkmark&\checkmark&\checkmark&\textbf{0.9383}& \textbf{0.9500}&\textbf{0.8998}&\textbf{0.9085}&\textbf{0.8943}&\textbf{0.9121}&\textbf{0.9232}&\textbf{0.9268}&\textbf{0.5828}&\textbf{0.6843}\\
			\bottomrule[0.8pt]
		\end{tabular}
	\end{center}
\end{table*}

\begin{figure*}[t]
	\centering
	\includegraphics[width=7.1in]{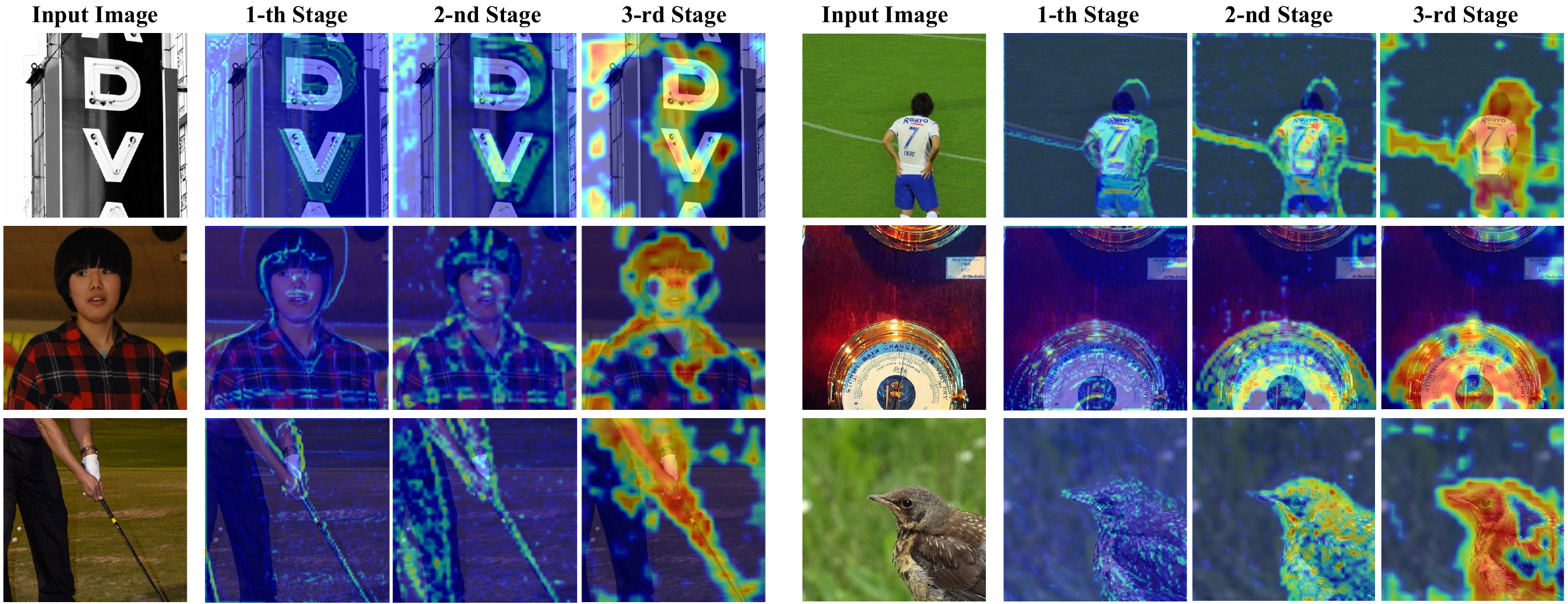}
	\caption{Activation maps of different stages of CvT obtained with the Grad-CAM method \cite{selvaraju2017grad}. It is observed that different stages of CvT extract different levels of image features, and later stages extract higher level features.}
	\label{stage}
  \vspace{-10pt}
\end{figure*}

\subsection{Cross-Database Validation}
We further conduct cross-database validations to test the generalizability of RichIQA, especially when faced with unseen data. 
Specifically, we train the method on an entire database and then test it on other databases to evaluate the generalizability of RichIQA and competing methods. 
Table \ref{Cross-Database_kon} presents the cross-database validation performances in predicting the MOS of image quality. 
Specifically, when trained on KonIQ-10K, RichIQA obtains 7.5\%, 4.9\%, 7.2\%, and 4.3\% SRCC gains on BID, LIVE Challenge, SPAQ, and FLIVE respectively.
When trained on BID, RichIQA obtains 12.9\%, 4.3\%, 0.1\%, 18.2\% SRCC gains on KonIQ-10K, LIVE Challenge, SPAQ, and FLIVE respectively.
The large and universal performance gains obtained  indicate that RichIQA has stronger generalizability compared to other methods.

 \begin{table*}[thp]
\renewcommand\arraystretch{1.3}
\tabcolsep=0.18cm
\begin{center}
\caption{
Impact of \textbf{Indirect and Direct Pathways} in the SLM Module on the Prediction Performance of RichIQA.
} \label{A-SLM}
\begin{tabular}{c||c|c|c|c |c|c |c|c |c|c}
  \toprule[0.8pt]
       \multicolumn{1}{c||}{} &\multicolumn{2}{c|}{KonIQ-10K }&\multicolumn{2}{c|}{BID}&\multicolumn{2}{c|}{LIVE Challenge}&\multicolumn{2}{c|}{SPAQ}&\multicolumn{2}{c}{FLIVE}\\ 
       \noalign{\vskip 0.5ex}\toprule[0.5pt]
         Method& SRCC$\uparrow$  & PLCC$\uparrow$  & SRCC$\uparrow$ & PLCC$\uparrow$ & SRCC$\uparrow$  & PLCC$\uparrow$  & SRCC$\uparrow$  & PLCC$\uparrow$  & SRCC$\uparrow$  & PLCC$\uparrow$ \\
         \noalign{\vskip 0.5ex}\toprule[0.8pt]
Method without direct pathway&\underline{0.9376}&\underline{0.9495}&0.8834&\underline{0.8977}&0.8907& \underline{0.9107}&0.9224& 0.9252&0.5740&0.6686\\
Method without indirect pathway&0.9361& 0.9479&0.8836&0.8885&\underline{0.8938}&0.9029&\underline{0.9232}& 0.9264&\underline{0.5795}&{0.6751}\\
Method without direct and indirect pathways&0.9363&0.9463&\underline{0.8671}&0.8946&0.8868& 0.9028&\textbf{0.9234}& \textbf{0.9273}&0.5736& \underline{0.6769}\\

 \noalign{\vskip 0.5ex}\cdashline{1-11}[2pt/2pt]\noalign{\vskip 1.0ex}

\textbf{RichIQA} &\textbf{0.9383}& \textbf{0.9500}&\textbf{0.8998}&\textbf{0.9085}&\textbf{0.8943}&\textbf{0.9121}&\underline{0.9232}&\underline{0.9268}&\textbf{0.5828}&\textbf{0.6843}\\
\bottomrule[0.8pt]
\end{tabular}
\end{center}
 \vspace{-10pt}
\end{table*}

  \begin{table*}[thp]
\renewcommand\arraystretch{1.2}
\tabcolsep=0.22cm
\begin{center}
\caption{ Impact of Using \textbf{Different Loss Functions} on the Prediction Performance of RichIQA. `\checkmark' Indicates that the Loss Function is Used. `\ding{55}' Indicates that the Loss Function is not Used.
} \label{A-Loss Function}
\begin{tabular}{c|c|c||c|c |c|c |c|c |c|c |c|c}
  \toprule[0.8pt]
       \multicolumn{3}{c||}{ Loss Function } &\multicolumn{2}{c|}{KonIQ-10K }&\multicolumn{2}{c|}{BID}&\multicolumn{2}{c|}{LIVE Challenge}&\multicolumn{2}{c|}{SPAQ}&\multicolumn{2}{c}{FLIVE}\\ 
       \noalign{\vskip 0.5ex}\toprule[0.5pt]
        $\mathrm{EMDLoss}$& $\mathrm{L1Loss}$& $\mathrm{ESDLoss}$ & SRCC$\uparrow$  & PLCC$\uparrow$  & SRCC$\uparrow$  & PLCC$\uparrow$ & SRCC$\uparrow$  & PLCC$\uparrow$  & SRCC$\uparrow$  & PLCC$\uparrow$  & SRCC$\uparrow$  & PLCC$\uparrow$ \\
        \noalign{\vskip 0.5ex}\toprule[0.8pt]
\checkmark&\ding{55}&\ding{55}&0.9364& 0.9473& \underline{0.8827}& \underline{0.9011}&\underline{0.8874}& \underline{0.9062}&0.9227&0.9251&0.5818&0.6744\\
\ding{55}&\checkmark&\ding{55}&0.9352& 0.9468&0.8808&0.8962&0.8832& 0.8963&\underline{0.9229}& \underline{0.9255}&0.5802&0.6820\\
\ding{55}&\ding{55}&\checkmark&0.2091& 0.2264&0.3433&0.3652&0.2816& 0.2840&0.4583&0.4693&0.1930   &       0.2042\\
\checkmark&\checkmark&\ding{55}&\underline{0.9377}&\underline{0.9489}&0.8814&0.9001&0.8857& 0.8989& 0.9225& 0.9241 &0.5806&{0.6824}\\
\checkmark&\ding{55}&\checkmark&0.9357&0.9472&0.8794&0.8954&0.8426& 0.8508&0.9185& 0.9217&0.5811&0.6606\\
\ding{55}&\checkmark&\checkmark&0.9365& 0.9461&0.8790&0.8922&0.8869& 0.9054&0.9225& 0.9264&\underline{0.5819}&\underline{0.6830}\\

 \noalign{\vskip 0.5ex}\cdashline{1-13}[2pt/2pt]\noalign{\vskip 1.0ex}

\checkmark&\checkmark&\checkmark&\textbf{0.9383}& \textbf{0.9500}&\textbf{0.8998}&\textbf{0.9085}&\textbf{0.8943}&\textbf{0.9121}&\textbf{0.9232}&\textbf{0.9268}&\textbf{0.5828}&\textbf{0.6843}\\
\bottomrule[0.8pt]
\end{tabular}
\end{center}
 \vspace{-10pt}
\end{table*}

\begin{figure*}[t]
	\centering
	\subfigure[KonIQ-10K] {
		\begin{minipage}[t]{0.24\linewidth}
			\centering
			\includegraphics[width=1\textwidth,trim=20 220 60 220, clip]{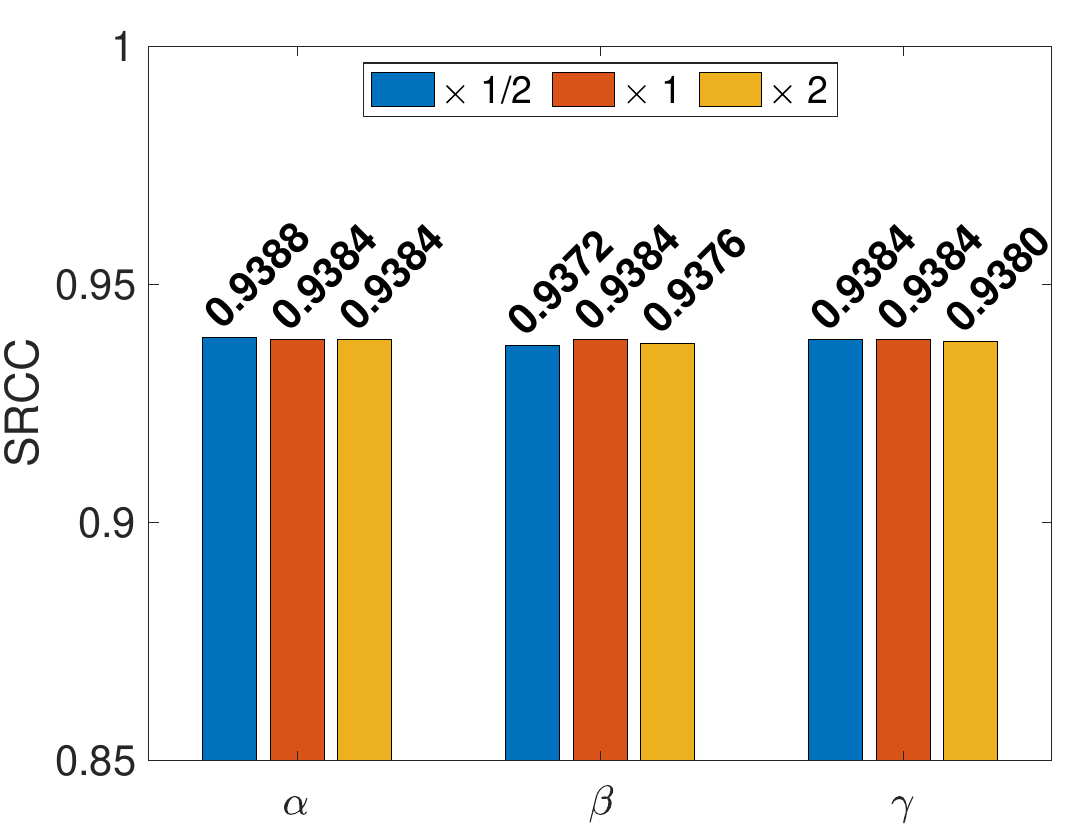}
		\end{minipage}\hfill
	}%
	\subfigure[BID] {
		\begin{minipage}[t]{0.24\linewidth}
			\centering
			\includegraphics[width=1\textwidth,trim=20 220 60 220, clip]{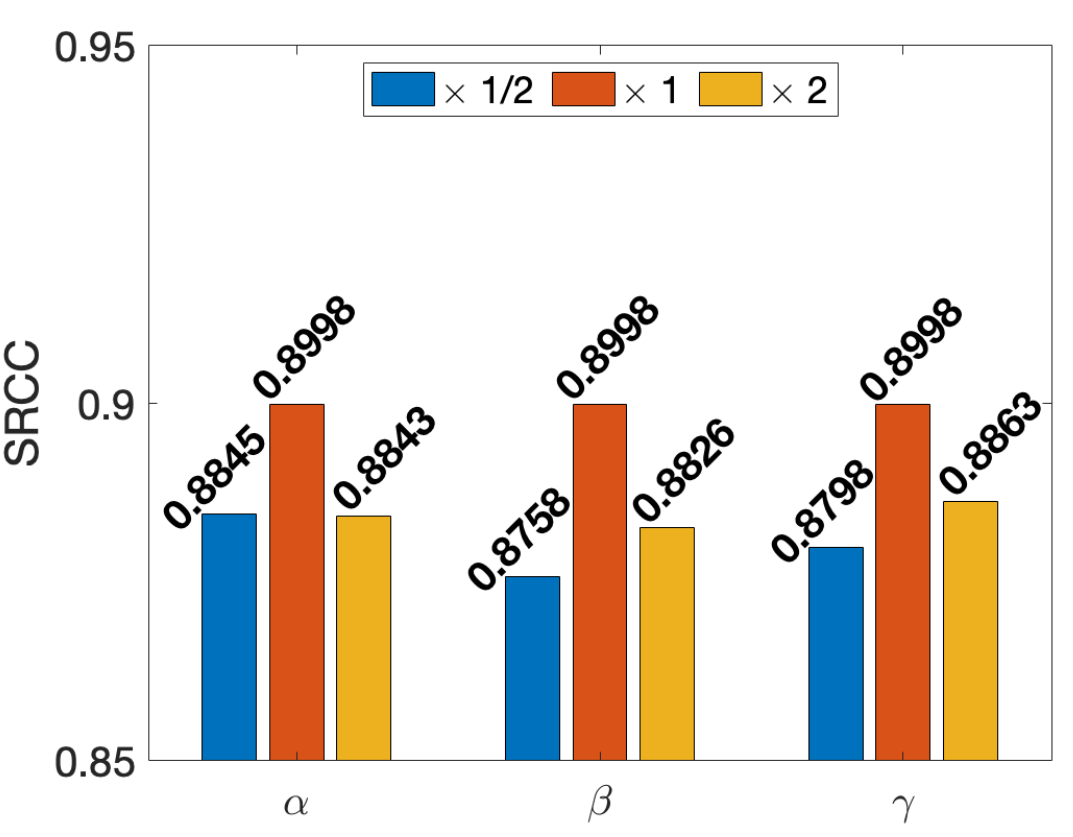}  
		\end{minipage}\hfill
	}%
	\subfigure[LIVE Challenge] {
		\begin{minipage}[t]{0.24\linewidth}
			\centering
			\includegraphics[width=1\textwidth,trim=20 220 60 220, clip]{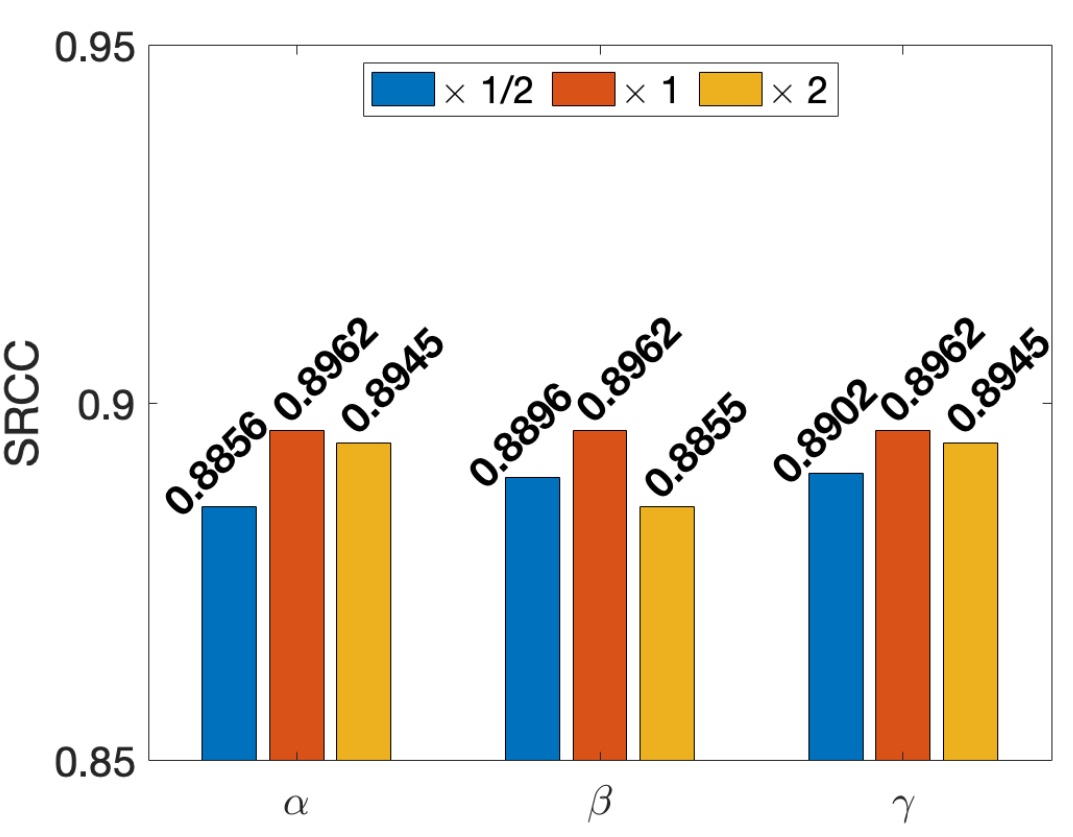}
		\end{minipage}
	}%
	\subfigure[SPAQ] {
		\begin{minipage}[t]{0.24\linewidth}
			\centering
			\includegraphics[width=1\textwidth,trim=20 220 60 220, clip]{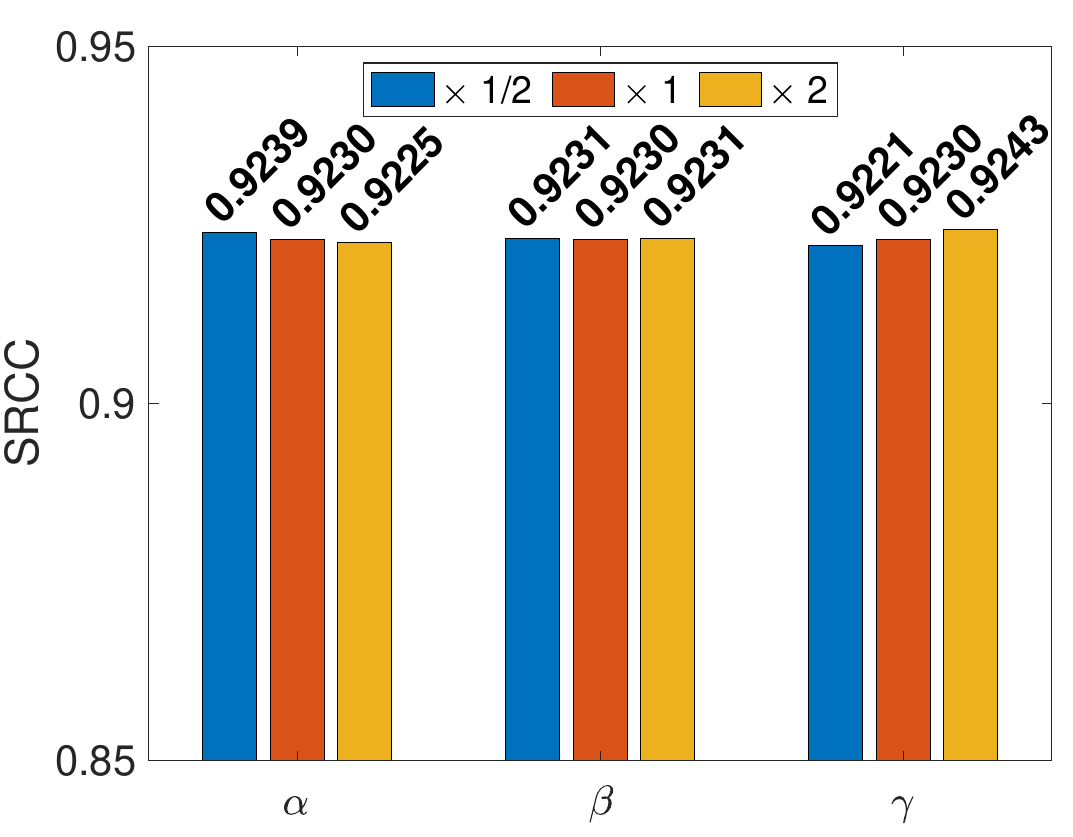}
		\end{minipage}
	}%
	\centering
	\caption{Impact of doubling or halving $\alpha$, $\beta$, and $\gamma$ on the prediction performance of RichIQA on KonIQ-10K, BID, LIVE Challenge, and SPAQ databases.}
	\label{BF}
  \vspace{-10pt}
\end{figure*}

\subsection{Ablation Experiment}
\subsubsection{Ablation with Different Stages of CvT}
We extract image features from three stages of CvT in this paper. We test the impact of the image features extracted from these three stages on the prediction performance of RichIQA. The results are shown in Table \ref{CvT}, from which it can be seen that using all three stages of CvT to extract image features can improve the prediction performance, and the lack of image features extracted from any stage can reduce the prediction performance. This indicates that the multi-level feature extraction paradigm in RichIQA is beneficial for image quality modeling. 

We also Grad-CAM \cite{selvaraju2017grad} to  compute the activation maps of different stages of CvT to visualize how each stage of the CvT influences the predictions made by RichIQA. 
Fig. \ref{stage} shows the activation maps of different stages of CvT obtained for six images using the Grad-CAM method.
From the activation maps, we can see that different stages of CvT extract different levels of image features. Specifically, the first stage of CvT primarily extracts low-level image features, such as edges and contours. 
In the second stage, the activation maps begin to show more details, indicating that RichIQA is focusing on more low-level features of the image, such as shapes and local structures.
In the third stage, the activation maps become more focused, demonstrating that RichIQA can extract high-level semantic features of the image, such as specific objects. 
From the activation maps and the quantitative ablation results, we can conclude that making full use of visual information from low-level to high-level can benefit visual quality assessment of in the wild images.

\subsubsection{Ablation with the SLM Module}
This paper proposes a SLM module to simulate the analysis and decision-making process of IQA in the human brain. The SLM module has two pathways through which the brain processes image features: a direct pathway and an indirect pathway. The direct pathway feeds the attention features of the image directly into the quality prediction module. The indirect pathway inputs the image features processed by the long-term memory mechanism into the quality prediction module. To validate these two pathways, three comparison methods are designed:
\begin{itemize}
	\item The first method excludes the direct pathway.
	\item The second method excludes the indirect pathway.
	\item The third method excludes both direct and indirect pathways, which means that there is no SLM module.
\end{itemize}

Table \ref{A-SLM} lists the performances of these three comparison methods and RichIQA in predicting MOS. 
For all databases, the exclusion of indirect or direct pathways in the SLM module can lead to a decrease in the performance. For most databases, such as KonIQ-10K, BID, LIVE Challenge, and FLIVE, excluding the SLM module can result in a significant decrease in the prediction performance.
In conclusion, the SLM module can improve the prediction performance of RichIQA on most in the wild IQA databases, and the indirect pathway and direct pathway in the SLM module are both essential.

\subsubsection{Ablation with Loss Functions}
We combine three loss functions ($\mathrm{EMDLoss}$, $\mathrm{L1Loss}$, and $\mathrm{ESDLoss}$) to jointly train the network. We test the impact of different loss functions on the performance of RichIQA, and list the results in Table \ref{A-Loss Function}. 
It is observed that RichIQA can achieve good prediction results when $\mathrm{EMDLoss}$ or $\mathrm{L1Loss}$, or any two of $\mathrm{EMDLoss}$, $\mathrm{L1Loss}$, and $\mathrm{ESDLoss}$, are used to train the network, and it has the best performance when the network is trained using all three loss functions simultaneously.

Additionally, we find that the loss function used by the second-best performing method is not uniform across different databases. For example, on KonIQ-10K, using both $\mathrm{EMDLoss}$ and $\mathrm{L1Loss}$ results in the second-best performance. On BID and LIVE Challenge, using $\mathrm{EMDLoss}$ results in the second-best performance. On SPAQ, using only $\mathrm{L1Loss}$ for training is sufficient to achieve the second-best performance. On FLIVE, using both $\mathrm{L1Loss}$ and $\mathrm{ESDLoss}$ results in the second-best performance. However, using $\mathrm{EMDLoss}$, $\mathrm{L1Loss}$ and $\mathrm{ESDLoss}$ simultaneously achieves the best performance across all databases. 
These findings highlight the feasibility and superiority of the proposed multi-label training strategy.

\subsubsection{Ablation with Balance Factors of Loss Functions}
We use three factors, $\alpha$, $\beta$, and $\gamma$, to balance the three loss functions ($\mathrm{EMDLoss}$, $\mathrm{L1Loss}$, and $\mathrm{ESDLoss}$) in a ratio of $200:10:1$. In this section, we test the impact of doubling or halving $\alpha$, $\beta$, and $\gamma$ on the performance on the KonIQ-10K, BID, LIVE Challenge, and SPAQ databases, respectively. The results are shown in Fig. \ref{BF}, from which it can be observed that on KonIQ-10K and SPAQ databases, the changes in the values of $\alpha$, $\beta$, and $\gamma$ have little effect on the performance of RichIQA. On BID and LIVE Challenge databases, any changes in $\alpha$, $\beta$, and $\gamma$ would decrease the  performance of RichIQA. Therefore, it is reasonable to use $\alpha$, $\beta$, and $\gamma$ with a ratio of $200:10:1$ to balance the loss function in this paper.

\begin{figure}[t]
	\centering
	\includegraphics[width=3.in]{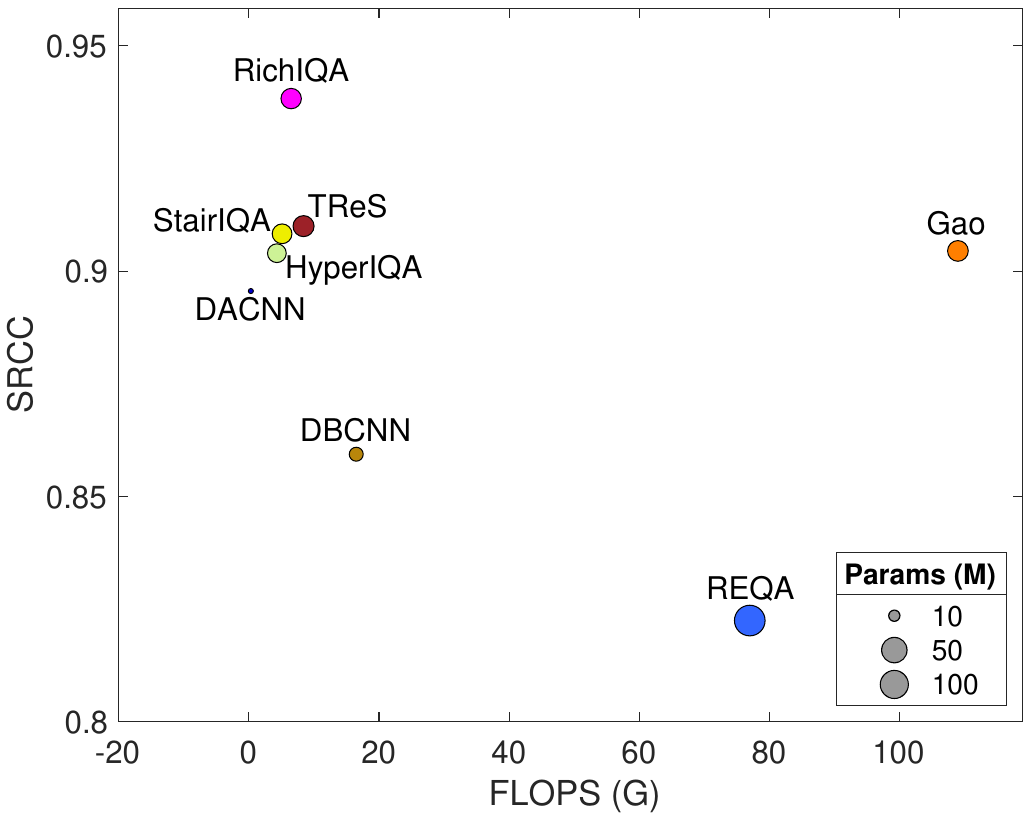}
	\caption{Comparison of FLOPS, network parameters, and prediction performances (on KonIQ-10K) between RichIQA and competing methods.
		Note that the input image size is $224 \times 224$. The number of network parameters is represented by the radius of the circle. 
	}
	\label{cc}
  \vspace{-10pt}
\end{figure}

\subsection{Computational Complexity}
Fig. \ref{cc} illustrates the computational complexity comparisons of RichIQA and other competing methods. We mainly compare the computational complexity of different methods based on the floating point operations per second (FLOPS) and network parameters, and the performance is tested on the KonIQ-10K database. 
From the figure, we can see that RichIQA, HyperIQA, StairIQA, DACNN, and TReS have relatively low computational complexity. The FLOPS of these methods is less than 15G, and the number of network parameters does not exceed 50M. Although the computational complexities of these methods are close, our proposed method has the best prediction performance. In summary, RichIQA can achieve the best balance between prediction performance and computational complexity.

\section{Conclusion}\label{sec:conclusion}
Aiming at NR IQA for ``images in the wild'' where human subjects may have more diverse perceptions, we propose an innovative NR IQA model RichIQA, which on one hand extracts multi-level features with a multi-stage CvT and performs deeper feature modeling with a short- and long-term memory module, and on the other exploits rich subjective quality rating information including MOS, SOS and DOS for model training. 
Different from most of the existing IQA models which only predict a single quality score, RichIQA predicts a quality distribution from which the mean image quality could also be predicted. 
Owing to the powerful feature extraction and modeling ability of the network as well as the explored rich subjective quality information beyond MOS, RichIQA consistently outperforms the state-of-the-art NR IQA models on five mainstream in the wild IQA databases. 
Large margin leads obtained in the cross-database validation also verify that RichIQA has good generalizability. 
Moreover, RichIQA achieves the best performances without increasing complexity too much. 
Though developed for NR IQA in the wild, the same methodology of exploring rich subjective quality information beyond MOS can be generalized to other perceptual quality assessment scenarios in the future. 

%


\ifCLASSOPTIONcaptionsoff
  \newpage
\fi



\bibliographystyle{IEEEtran}

\bibliography{IEEEabrv,refs}

\begin{IEEEbiography}[{\includegraphics[width=1in,height=1.25in,clip,keepaspectratio]{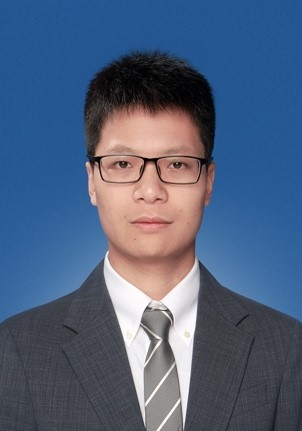}}]{Xiongkuo Min} (Member, IEEE)
received the B.E. degree from Wuhan University, Wuhan, China, in 2013, and the Ph.D. degree from Shanghai Jiao Tong University, Shanghai, China, in 2018, where he is currently a tenure-track Associate Professor with the Institute of Image Communication and Network Engineering. From Jan. 2016 to Jan. 2017, he was a visiting student at University of Waterloo. From Jun. 2018 to Sept. 2021, he was a Postdoc at Shanghai Jiao Tong University. From Jan. 2019 to Jan. 2021, he was a visiting Postdoc at The University of Texas at Austin and the University of Macau. He received the Best Paper Runner-up Award of IEEE Transactions on Multimedia in 2021, the Best Student Paper Award of IEEE International Conference on Multimedia and Expo (ICME) in 2016, and the excellent Ph.D. thesis award from the Chinese Institute of Electronics (CIE) in 2020. His research interests include image/video/audio quality assessment, quality of experience, visual attention modeling, extended reality, and multimodal signal processing. 
\end{IEEEbiography}
\begin{IEEEbiography}[{\includegraphics[width=1in,height=1.25in,clip,keepaspectratio]{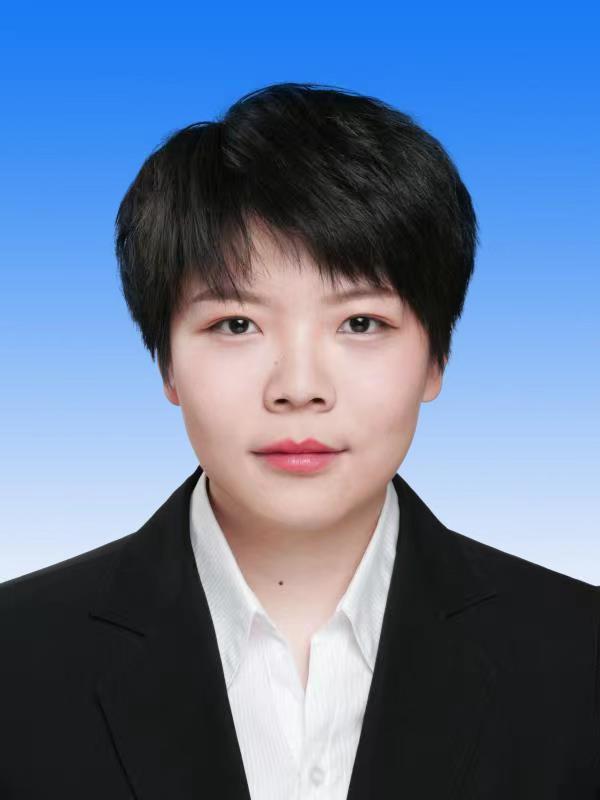}}]{Yixuan Gao}
received the B.E. degree from the Harbin Institute of Technology, Weihai, China, in 2020. She is currently working toward a Ph.D. degree with the Institute of Image Communication and Network Engineering, Shanghai Jiao Tong University, Shanghai, China. Her current research interest is in image quality assessment.
\end{IEEEbiography}
\begin{IEEEbiography}[{\includegraphics[width=1in,height=1.25in,clip,keepaspectratio]{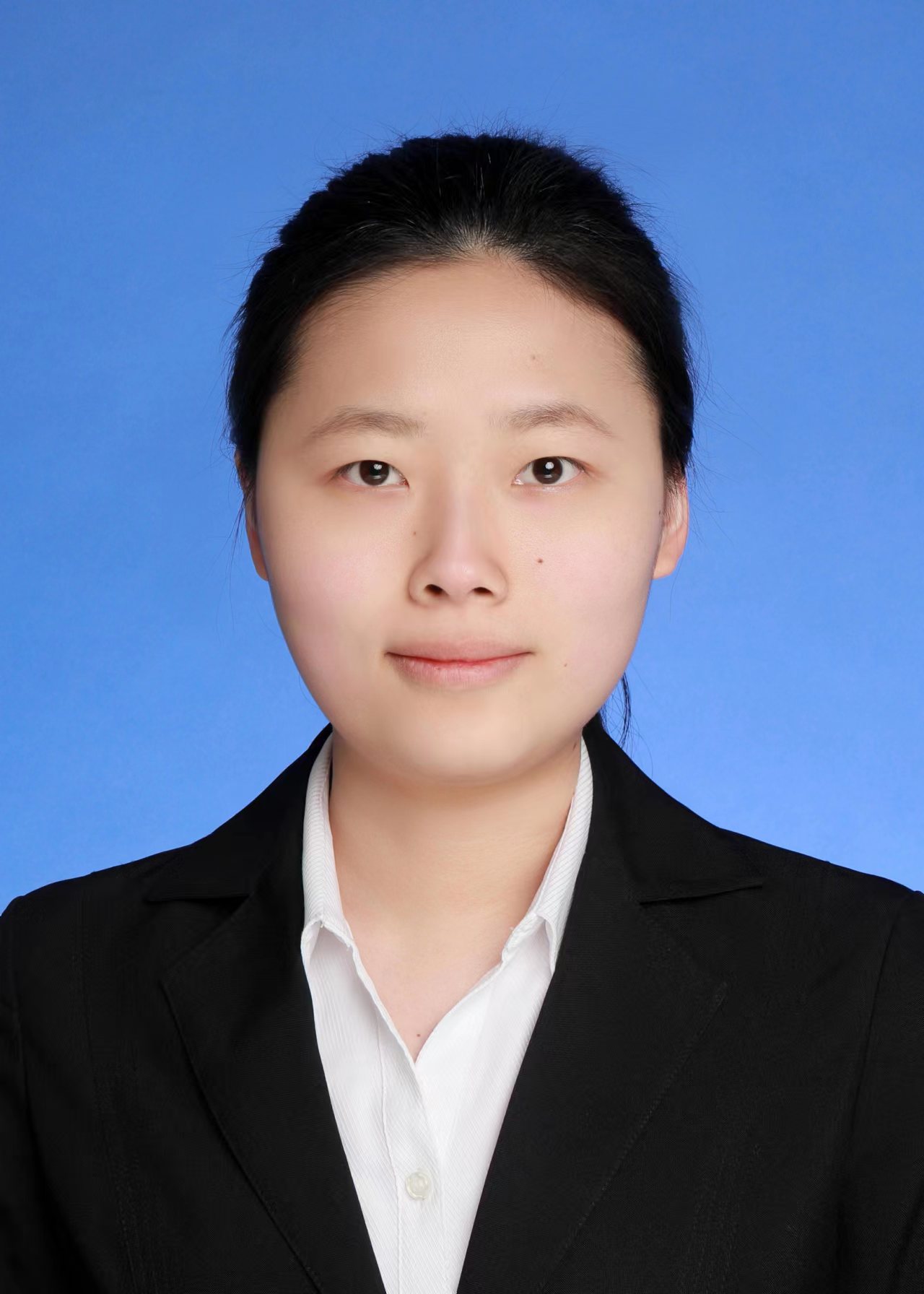}}]{Yuqin Cao}
received the B.E. degree from Shanghai Jiao Tong University, Shanghai, China, in 2021, where she is currently pursuing the Ph.D. degree with the Institute of Image Communication and Network Engineering. Her current research interests include audio-visual quality assessment.
\end{IEEEbiography}

\begin{IEEEbiography}[{\includegraphics[width=1in,height=1.25in,clip,keepaspectratio]{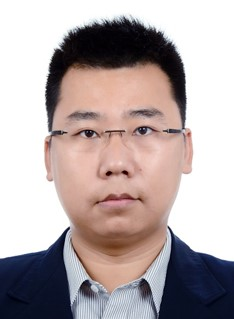}}]{Guangtao Zhai} (Senior Member, IEEE)
 received the B.E. and M.E. degrees from Shandong University, Shandong, China, in 2001 and 2004, respectively, and the Ph.D. degree from Shanghai Jiao Tong University, Shanghai, China, in 2009. From 2008 to 2009, he was a Visiting Student with the Department of Electrical and Computer Engineering, McMaster University, Hamilton, ON, Canada, where he was a Postdoctoral Fellow from 2010 to 2012. From 2012 to 2013, he was a Humboldt Research Fellow with the Institute of Multimedia Communication and
Signal Processing, Friedrich Alexander University of Erlangen-Nuremberg, Germany. He is currently a Professor at the Department of Electronics Engineering, Shanghai Jiao Tong University. He has published more than 100 journal articles on topics, including visual information acquisition, image processing, and perceptual signal processing. He is a member of IEEE CAS VSPC TC and MSA TC. He has received multiple international and domestic research awards, including the National Ph.D. thesis awards 2012, the Best Student Paper Award of IEEE ICME 2016, the Best Student Paper Award of PCS 2015, the Best Paper Award of IEEE TRANSACTIONS ON MULTIMEDIA 2018, Saliency360! Grand Challenge of ICME 2018, the Best Paper Award of IEEE MMC Workshop 2019, the Best Paper Award of IEEE CVPR DynaVis Workshop 2020, the Eastern Scholar and Dawn program professorship of Shanghai, the NSFC Excellent Young Researcher Program, and the National Top Young Researcher Award. He will be serving as the Editor-in-Chief for Displays (Elsevier) and is on the Editorial Board for Digital Signal Processing (Elsevier) and Science China: Information Science.
\end{IEEEbiography}

\begin{IEEEbiography}[{\includegraphics[width=1in,height=1.25in,clip,keepaspectratio]{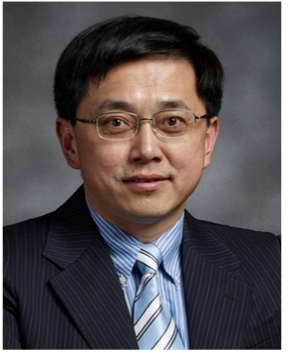}}]{Wenjun Zhang}
 (Fellow, IEEE) received the B.S., M.S. and Ph.D. degrees in electronic engineering from Shanghai Jiao Tong University, Shanghai, Chi- na, in 1984, 1987 and 1989, respectively. From 1990 to 1993, Heworked as a post-doctoral fellow at Philips Kommunikation Industrie AG in Nuremberg, Germany, where he was actively involved in devel- oping HDMAC system. He joined the Faculty of Shanghai Jiao Tong University in 1993 and became a full professor in the Department of Electronic Engineering in 1995. As the national HDTV TEEG project leader, he successfully developed the first Chinese HDTV prototype system in 1998. He was one of the main contributors to the Chinese Digital Television Terrestrial Broadcasting Standard issued in 2006 and is leading team in designing the next generation of broadcast television system in China from 2011. He holds more than 40 patents and published more than 90 papers in international journals and conferences. His main research interests include digital video coding and transmission, multimedia semantic processing and intelligent video surveillance. He is a Chief Scientist of the Chinese National Engineering Research Centre of Digital Television (NERC-DTV), an industry/government consortium in DTV technology research and standardization and the Chair of Future of Broadcast Television Initiative (FOBTV) Technical Committee.
\end{IEEEbiography}

\begin{IEEEbiography}[{\includegraphics[width=1in,height=1.25in,clip,keepaspectratio]{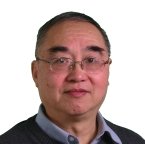}}]{Huifang Sun}
  (Fellow, IEEE) received the
B.Sc. degree from the Harbin Military Engineering
Institute, Harbin, China, and the Ph.D. degree from
the University of Ottawa, Ottawa, ON, Canada.
He was an Associate Professor with Fairleigh
Dickinson University, Teaneck, NJ, USA, in 1990.
He joined Sarnoff Corporation, Princeton, NJ, USA,
in 1990, as a member of the Technical Staff. He was
promoted to Technology Leader with Digital Video
Communication, Seattle, WA, USA. In 1995, he
joined Mitsubishi Electric Research Laboratories,
Cambridge, MA, USA, where he was promoted to Vice President and Deputy
Director in 2003 and is currently a Fellow. He has co-authored two books
and authored over 150 journal and conference papers. He holds over 61 U.S.
patents.
Dr. Sun received the Technical Achievement Award for optimization and
specification of the Grand Alliance HDTV video compression algorithm at the
Sarnoff Laboratory in 1994. He also received the best paper award of the IEEE
TRANSACTIONS ON CONSUMER ELECTRONICS in 1992, the International
Conference on Consumer Electronics in 1996, and the IEEE TRANSACTIONS
ON CIRCUITS AND SYSTEMS FOR VIDEO TECHNOLOGY in 2003. He was an
Associate Editor of the IEEE TRANSACTIONS ON CIRCUITS AND SYSTEMS
FOR VIDEO TECHNOLOGY and the Chair of the Visual Processing Technical
Committee of the IEEE Circuits and System Society
\end{IEEEbiography}

\begin{IEEEbiography}[{\includegraphics[width=1in,height=1.25in,clip,keepaspectratio]{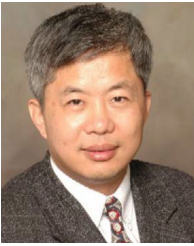}}]{Chang Wen Chen}
(Fellow, IEEE) received the B.S. degree
from the University of Science and Technology of
China, Hefei, China, in 1983, the M.S.E.E. degree
from the University of Southern California, Los
Angeles, CA, USA, in 1986, and the Ph.D. degree
from the University of Illinois at Urbana-
Champaign, Champaign, IL, USA, in 1992.
He was with the Faculty of Electrical and
Computer Engineering, University of Rochester,
Rochester, NY, USA, from 1992 to 1996, and the
Faculty of Electrical and Computer Engineering,
University of Missouri-Columbia, Columbia, MO, USA, from 1996 to 2003.
He was the Allen Henry Endow Chair Professor with the Florida Institute
of Technology, Melbourne, FL, USA, from 2003 to 2007. He is currently
a Professor of Computer Science and Engineering with the University at
Buffalo, The State University of New York, Buffalo, NY, USA.
\end{IEEEbiography}

%
%
%

%




\end{document}